\documentclass[11pt]{article}
 
\usepackage[top=1in,bottom=1in,left=1in,right=1in]{geometry}
\usepackage[numbers]{natbib}

\usepackage{algorithm}
\usepackage{algorithmicx}
\usepackage{amsmath,amssymb,amsfonts}
\usepackage{graphicx,color}
\usepackage{url}
\usepackage{subfiles}
\usepackage{booktabs}
\usepackage{mathtools}
\usepackage{microtype}
\usepackage[english]{babel}
\usepackage{caption}
\usepackage[toc,page,title,titletoc]{appendix}

\usepackage[dvipsnames]{xcolor}
\usepackage{tikz, pgfplots, pgfplotstable, subcaption}
\usepackage{hyperref}       
\hypersetup{
  colorlinks=true,
  linkcolor=black,
  urlcolor=black,
  citecolor=black
}
\usepackage[noend]{algpseudocode}
\usepackage{setspace}
\usepackage{yhmath}
\usepackage[figurename=Fig.,font={small,stretch=0.84}]{caption}
 
\usepackage[numbers]{natbib}
\usepackage{authblk}
\usepackage{cleveref}
\crefname{section}{Section}{Sections}
\crefname{figure}{Figure}{Figures}
\crefname{mythm}{Theorem}{Theorems}
\crefname{mylem}{Lemma}{Lemmas}
\crefname{myrem}{Remark}{Remarks}
\crefname{appendix}{Appendix}{Appendicies}
\crefname{myprop}{Proposition}{Propositions}
\crefname{equation}{Equation}{Equations}
\crefname{algorithm}{Algorithm}{Algorithms}

\usetikzlibrary{arrows.meta, calc, topaths, positioning, automata}
\pgfplotsset{compat=1.16}
\newenvironment{customlegend}[1][]{%
\begingroup
\csname pgfplots@init@cleared@structures\endcsname
\pgfplotsset{#1}%
}{%
\csname pgfplots@createlegend\endcsname
\endgroup
}%
\def\addlegendimage{\csname pgfplots@addlegendimage\endcsname}
\def\BibTeX{{\rm B\kern-.05em{\sc i\kern-.025em b}\kern-.08em
    T\kern-.1667em\lower.7ex\hbox{E}\kern-.125emX}}

\newcommand{\prob}[1]{\mathbb{P}( #1 )}
\newcommand{\Prob}[1]{\mathbb{P}\left( #1 \right)}

\newcommand{\norm}[1]{ \|   #1  \| }
\newcommand{\ind}[1]{ \mathbf{I}_{\{#1 \}} }
\newcommand{\argmax}[0]{ \mathop{\arg\max}  }
\newcommand{\argmin}[0]{ \mathop{\arg\min}  }

\newtheorem{mythm}{\bf{Theorem}}[section]
\newtheorem{mylem}{\bf{Lemma}}[section]
\newtheorem{myprop}{\bf{Proposition}}[section]
\newtheorem{mydef}{\bf{Definition}}[section]
\newtheorem{mycol}{\bf{Corollary}}[section]

\newtheorem{myasp}{\bf{Assumption}}[section]
\newenvironment{myproof}{ {\noindent\it Proof.\ }}{\hfill $\square$\par}

\algrenewcommand{\algorithmiccomment}[1]{\hfill// #1}

\usepackage{multirow}

\title{
  Parameter-Adaptive Dynamic Pricing
}

\usepackage{authblk}
\author[*]{Xueping Gong}
\author[$\dagger$]{Jiheng Zhang}

\affil[*$\dagger$]{Department of Industrial Engineering and Decision Analytics}
\affil[*$\dagger$]{The Hong Kong University of Science and Technology}

\date{}

\begin{document}
\maketitle

\begin{abstract}
  Dynamic pricing is crucial in sectors like e-commerce and transportation, 
  balancing exploration of demand patterns and exploitation of pricing strategies. 
  Existing methods often require precise knowledge of the demand function,
  e.g., the H{\"o}lder smoothness level and Lipschitz constant, 
  limiting practical utility. This paper introduces an adaptive approach to address these challenges without prior parameter knowledge. By partitioning the demand function's domain and employing a linear bandit structure, we develop an algorithm that manages regret efficiently, enhancing flexibility and practicality. Our Parameter-Adaptive Dynamic Pricing (PADP) algorithm outperforms existing methods, offering improved regret bounds and extensions for contextual information. Numerical experiments validate our approach, demonstrating its superiority in handling unknown demand parameters.
\end{abstract}

\section{Introduction}

Dynamic pricing, a technique involving the real-time modulation of prices in response to changing market dynamics, 
has emerged as a crucial strategy in sectors like e-commerce and transportation. 
A successful dynamic pricing framework must strike a delicate balance between exploration—by learning demand patterns across different price points—and exploitation—by fine-tuning prices based on observed data on pricing and demand.
This field has garnered considerable scholarly interest due to its practical implications and its pivotal role in revenue optimization.
For an in-depth exploration of the existing literature on dynamic pricing, 
readers are encouraged to consult \citep{DP_review}. 
Additionally, for valuable insights into specific applications of dynamic pricing strategies, 
\citet{DP_application} offer a comprehensive perspective that can further enrich our understanding of this dynamic field.

An inherent challenge prevalent in existing methodologies lies in the requirement for precise knowledge of the H{\"o}lder smoothness level $\beta$ and the Lipschitz constant $L$ governing the demand function.
These assumptions often fail to align with the intricacies of real-world scenarios, thereby limiting the practical utility of these algorithms.
In response to this limitation, our research distinguishes itself by venturing into unexplored terrain concerning adaptability within dynamic pricing systems.
Specifically, we tackle the critical issue of adaptation when confronted with unknown values of the H{\"o}lder smoothness level $\beta$ and the Lipschitz constant $L$, a challenge that has remained largely unaddressed in prior works.

To address this challenge effectively, we introduce a novel adaptive methodology designed to navigate the inherent uncertainties surrounding key parameters. 
Initially, we partition the domain of the demand function into equidistant intervals, a foundational step in our approach.
This discretization not only gives rise to a linear bandit structure but also facilitates the formulation of upper confidence bounds adept at accommodating biases arising from approximation errors. 
Building upon the unique characteristics of this linear bandit framework, we implement a stratified data partitioning technique to efficiently regulate regret in each iteration.
The adaptive nature of this methodology obviates the necessity for a Lipschitz constant, 
thereby amplifying the flexibility and practicality of our strategy. 
In handling the smoothness parameter, we introduce an estimation method under a similarity condition, 
enabling seamless adaptation to unknown levels of smoothness.
This method eliminates the need for the smoothness parameter, 
enhancing the flexibility and practicality of our approach.
Lastly, through meticulous consideration, we determine an optimal number of discretization intervals to attain minimax optimality for this intricate problem.

Our contributions in this paper can be succinctly summarized as follows:
\begin{itemize}
    \item Innovative adaptive algorithm: We introduce the Parameter-Adaptive Dynamic Pricing (PADP) algorithm, 
    employing the layered data partitioning technique to eliminate the dependence on unknown parameters of the demand function.
    Moreover, our algorithm is designed to adapt to model misspecification, enhancing its robustness for real-world applications.
    \item Improved regret bound:  Our proposed algorithm outperforms existing works such as \cite{PlinearDP} and \cite{Multimodal_DP} by achieving a superior regret upper bound. 
    We address limitations present in the regret bound of \cite{Multimodal_DP}, 
    enhancing scalability in terms of price bounds and improving the order concerning the smoothness level $\beta$.
    \item Extension: We present additional extensions, including the incorporation of linear contextual information. 
    By comparing with the established linear contextual effect model as detailed in \citep{PlinearDP}, 
    our method excels in enhancing the leading order with respect to contextual dimensions. 
    Additionally, our method can recover the order concerning the Lipschitz constant in \citep{PlinearDP} when this parameter is provided in advance.
    \item Solid numerical experiments: Through a series of comprehensive numerical experiments, 
    we underscore the necessity of understanding the parameters of the demand function. 
    In the absence of this knowledge, previous methods either falter or exhibit degraded performance. 
    This empirical evidence underscores the superiority of our method.
\end{itemize}

\subsection{Related work}

\textbf{Nonparametric dynamic pricing.}
Dynamic pricing has been an active area of research, 
driven by advancements in data technology and the increasing availability of customer information.
Initial research focused on non-contextual dynamic pricing without covariates \citep{non_DP_linear,DP_finiteV}.
For example, \citet{Multimodal_DP} employed the UCB approach with local-bin approximations, achieving an $\tilde{\mathcal{O}}(T^{(m+1)/(2m+1)})$ regret for $m$-th smooth demand functions and establishing a matching lower bound. 
\citet{PlinearDP} extend this model into the one with additive linear contextual effect. 
\citet{PlinearDP} devise a different learning algorithm based on the biased linear contextual bandit borrowed from \citet{Multimodal_DP} 
and their new idea of being more optimistic to chase the context-dependent optimal price. 
They establish an instance-dependent bounds and a matching lower bound. 

\textbf{Contextual dynamic pricing.}
The realm of context-based dynamic pricing has been extensively explored in the literature, 
as evidenced by studies such as \cite{non_DP_linear}, \cite{OfflineDP}, \cite{DP_generalV}, and \cite{PersonalizedDP}. 
Within this domain, the binary choice model, exemplified by works like \cite{DP_parametricF}, \cite{DP_ambiguity_set}, \cite{DP_cox}, \cite{LP_LV}, \cite{d_free_DP}, \cite{Explore_UCB}, and \cite{DP_Fm}, 
assumes that every customer's purchase decision follows a Bernoulli distribution.
However, our model exhibits a broader scope, 
accommodating non-binary demand scenarios and encompassing the binary case as a special instance. 
For a comprehensive overview of recent developments and related works, 
we direct the reader to \cite{PlinearDP} and \cite{Multimodal_DP}. 
Compared with these work, our approach reduces the order of smoothness level in the regret upper bound and obviates the necessity of a Lipschitz constant in the algorithmic implementation.
Achieving such outcomes necessitates meticulous parameter selection within the algorithm and a sophisticated algorithmic design.

\textbf{Contextual bandit.} The most studied model in contextual bandit is the linear model (see, e.g., \citep{banditBook,CBwithRegressionOracles,improvedUCB,misspecified_lin,supLinUCB}), 
where the expected reward is a linear combination of contexts. 
The algorithms developed in these works are mostly built upon the celebrated idea of the OFU principle. 
Specifically, our approach relates to the concept of misspecified linear bandits \citep{supLinUCB,misspecified_lin} within the context of dynamic pricing.
Unlike traditional bandit algorithms, dynamic pricing requires consideration of both estimation variance and the approximation errors.
Notably, we have discovered an intriguing finding that by leveraging the unique structure of the dynamic pricing problem, 
we can achieve a more precise and improved regret bound compared to directly applying complex existing algorithms designed for misspecified linear bandits. 
This improvement is mainly attributed to the fact that the bias encountered in each round is consistent across all prices (actions).
This underscores the importance of leveraging the distinctive structure and attributes of the pricing context in order to achieve superior performance.
There are also a substantial amount of literature considering non-parametric reward feedback under
H{\"o}lder continuous assumption (see, e.g., \citep{Multimodal_DP,PlinearDP}). 

\textbf{Notations.} 
Throughout the paper, we use the following notations. For any positive integer $n$, we denote the set $\{1,2,\cdots,n\}$ as $[n]$. The cardinality of a set $A$ is denoted by $|A|$.
We use $\ind{E}$ to represent the indicator function of the event $E$. Specifically, $\ind{E}$ takes the value $1$ if $E$ happens, and $0$ otherwise.
For norms, we utilize the notation $\norm{\cdot}_p$ where $1\leq p\leq \infty$ to denote the $\ell_p$ norm.
Throughout the analysis, the notation $\tilde{\mathcal{O}}$ is employed to hide the dependence on absolute constants and logarithmic terms. It allows us to focus on the dominant behavior of the quantities involved.

\section{Basic Setting}
\label{sec: basic setting}

We consider a stylized dynamic pricing setting consisting of $T$ selling time periods, 
conveniently denoted as $t=1,2,\cdots,T$. 
At each time period $t$ before a potential customer comes, 
a retailer sets a price $p_t \in [p_{min},p_{max}]$, where $[p_{min},p_{max}]$ is a predetermined price range. 
The retailer then observes a randomized demand $d_t \in [0, 1]$ and collects a revenue of $r_t= p_td_t$. 
We constrain the realized demand at every single selling period $t$ to be at most one, 
which can be achieved by considering short arrival time periods such that at most one purchase will occur within each time period.

The randomly realized demands $\{d_t\}^T_{t=1}$ given advertised prices $\{p_t\}^T_{t=1}$ are governed by an underlying demand function $f : [p_{min}, p_{max}] \to [0, 1]$ such that at each time $t$, $d_t \in [0, 1]$ almost surely 
and $\mathbb{E}[d_t |p = p_t] = f (p_t)$. That is, the demand function $f (p)$ specifies the expected demand under posted prices. 
In this paper, the demand function $f$ is assumed to be unknown prior to the whole selling seasons and has to be learnt on the fly as the selling proceeds.
The regret at time $t$ is defined as the loss in reward resulting from setting the price $p_t$ compared to the optimal price $p^* = \argmax_{p\in [p_{min},p_{max}]} pf(p)$. 
The cumulative regret over the horizon of $T$ periods, denoted as $Reg(T)$, is given by the expression:
$$
Reg(T) = \sum_{t=1}^T \left(\max_{p \in [p_{min},p_{max}]} p f(p) - p_tf(p_t)\right).
$$
To evaluate the performance, we consider the expected cumulative regret $\mathbb{E}[Reg(T)]$, 
which takes into account the randomness of the potential randomness of the pricing policy.
The goal of our dynamic pricing policy is to decide the price $p_t$ at time $t$, 
by utilizing all historical data $\{(p_s, y_s), s \in [t-1]\}$, in order to minimize the expected cumulative regret.

To enable learning of the underlying demand function $f$, we place the following smoothness condition on $f$:
\begin{myasp}(Smoothness Condition)
    \label{asp: Smoothness}
    For some $\beta>1$, and $L>0$, the demand function $f : [p_{min}, p_{max}] \to [0, 1]$ is H{\"o}lder-continuous of exponent $\varpi(k)$ and Lipschitz constant $L$ on domain $[p_{min} , p_{max} ]$, 
    meaning that $f$ is $\varpi(k) $-times differentiable on $[p_{min}, p_{max}]$, and furthermore
    \begin{align*}
       & | f^{(\varpi(\beta))}(p) - f^{(\varpi(\beta))}(p')   | \leq L |p -p' |^{\beta -\varpi(\beta)  },  \forall p,p' \in [p_{min}, p_{max}].
    \end{align*}
\end{myasp}

It is essential to recognize that several widely accepted smoothness conditions in existing literature are included in \cref{asp: Smoothness}. 
For example, functions with $\beta=1$ belong to the class of Lipschitz continuous functions, as outlined in \citep{besbes2009dynamic}. 
When $\beta=2$, this encompasses all functions with bounded second-order derivatives, consistent with the assumptions in \citep{DP_inventory_constraint}. 
For any general integer $\beta$, \cref{asp: Smoothness} aligns with Assumption 1 in \citep{Multimodal_DP}.

It is important to note that, unlike various studies that assume concavity (e.g., \citep{DP_2}), we do not make this assumption here. 
Furthermore, this class of functions depends on the smoothness constant $L$, 
which intuitively reflects how closely $f$ can be approximated by a polynomial.
Since $f$ is $\varpi(k) $-times differentiable on the bounded interval $[p_{min}, p_{max}]$,
we can easily demonstrate that
$$
\sup_{ p\in [p_{min}, p_{max}] }  | f^{(k)}(p)  | < +\infty, \forall 0\leq k\leq \varpi(\beta).
$$
Compared with \citep{PlinearDP} which relies on the knowledge of the upper bounds of derivatives, 
our algorithms can be more practical and eliminate the need of this prior knowledge.

\section{Parameter Adaptive Dynamic Pricing}

In this section, we present an adaptive policy designed in the presence of unknown $L$. 
We also derive minimax regret upper bounds for our proposed algorithm as well as additional special cases (very smooth demands) with further tailored regret bounds.

\subsection{Algorithm design}

In this subsection, our method is presented in \cref{alg: PADP}.
The high-level idea of \cref{alg: PADP} is the following.
First, the algorithm partitions the price range $[p_{min}, p_{max}]$ into $N$ intervals.
We regard each interval as an action in bandit and uses the UCB technique to control the exploration and exploitation.
The UCB construction is inspired by linear contextual bandits,
using local polynomial approximates of the demand function $f$.

\begin{algorithm}[htp]
	\renewcommand{\algorithmicrequire}{\textbf{Input:}}
	\renewcommand{\algorithmicensure}{\textbf{Output:}}
    \algnewcommand{\LeftComment}[1]{\Statex \quad\quad  \(//\) \textit{#1}}
	\caption{Parameter Adaptive Dynamic Pricing}
	\label{alg: PADP}
	\begin{algorithmic}[1]
      \Require the discretization number $N$, the length $T$, the smoothness parameter $\beta$, the confidence parameter $\delta$ and the price bound $[p_{min},p_{max}]$
      \State Let $a_j = p_{min} + j\frac{(p_{max} - p_{min})}{N}$ for $j=0,1,\cdots,N$
      \State Let $\phi_j(p) = (1,p-a_j,\cdots,(p-a_j)^{\varpi (\beta)})^{\top}$
      \State Set $S = \lceil \log_2 \sqrt{ T} \rceil$, $\gamma = \sqrt{ \frac{1}{2}\ln(2NST/\delta) }$, $\Psi_{t}^s = \emptyset$ for $s\in [S]$
      \LeftComment{Initialization}
      \For{ round $t = 1,2,\cdots,N(1+\varpi(\beta))$ }
        \State Let $j=t \mod N $
        \State Set a distinct price $p_t$ in $[a_{j-1},a_j]$ and obtain the demand $d_t$
        \State Update the round collection $\Psi_{t+1}^{\sigma} = \Psi_{t}^{\sigma} \cup \{t\}$ for all $\sigma\in [S]$
      \EndFor
      \For{ round $t = N(1+\varpi(\beta))+1,N(1+\varpi(\beta))+2,\cdots,T$ }
         \State Let $s=1$ and $\mathcal{A}_{t,1} = \{j\in[N]| p_{\tau}\in [a_{j-1},a_j]  ,\tau \in [t-1] \}$
         \Repeat 
         \State Compute $\mathcal{D}_{t,s}^j = \{ \tau \in \Psi_t^s | p_{\tau}\in [a_{j-1},a_j] \}$ for all $j\in \mathcal{A}_{t,s}$
         \State For all $j\in\mathcal{A}_{t,s}$, compute the parameter 
         $
         \theta^j_{t,s} =  \argmin_{\theta }  \sum_{\tau \in \mathcal{D}_{t,s}^j }{ (d_{\tau} - \theta^\top  \phi_j(p_{\tau})  )^2   } 
         $
         \State For all $j\in\mathcal{A}_{t,s}$, compute the Gram matrix
         $
         \Lambda^j_{t,s} =   \sum_{\tau \in \mathcal{D}_{t,s}^j }  \phi_j(p_{\tau}) \phi_j(p_{\tau})^{\top}
         $
         \State Compute the upper confidence bound 
         $
         U^j_{t,s}(p) =  \phi_j(p )^{\top} \theta^j_{t,s} + \gamma \sqrt{ \phi_j(p )^{\top}  (\Lambda^j_{t,s})^{-1} \phi_j(p ) }
         $
         \If{ $   \sup_{ p \in [a_{j-1},a_j] } p \sqrt{ \phi_j(p )^{\top}  (\Lambda^j_{t,s})^{-1} \phi_j(p ) } \leq p_{max}/\sqrt{T} $ for all $ j \in \mathcal{A}_{t,s} $}
         \State Choose the price
         $
         p_t = \argmax_{j\in \mathcal{A}_{t,s}}\sup_{p \in [a_{j-1},a_j] } p U^j_{t,s}(p)
         $
         \State Update the round collection $\Psi_{t+1}^{\sigma} = \Psi_{t}^{\sigma}$ for all $\sigma\in [S]$
         \LeftComment{step (a)}
         \ElsIf{ $ \sup_{p\in [a_{j-1},a_j]}   p \sqrt{ \phi_j(p )^{\top}  (\Lambda^j_{t,s})^{-1} \phi_j(p ) } \leq p_{max} 2^{-s} $ for all $j \in \mathcal{A}_{t,s}$ }
         \State Let $\mathcal{A}_{t,s+1} = \{ j \in \mathcal{A}_{t,s}| \sup_{ p \in [a_{j-1},a_j] } p  U^j_{t,s}(p)  \geq  \max \limits_{  j' \in \mathcal{A}_{t,s} } \sup_{p\in [a_{j-1},a_j] } p U^{j'}_{t,s}(p)  -  p_{max} 2^{1-s}      \}$
         \State Let $s \leftarrow s+1  $
         \LeftComment{step (b)}
         \Else 
         \State Choose $j_t$ such that $\sup_{p\in [a_{j_{t}-1},a_{j_t}]} p \gamma \sqrt{ \phi_j(p )^{\top}  (\Lambda^j_{t,s})^{-1} \phi_j(p ) } >   p_{max}2^{-s}$
         \State Choose the price 
         $
         p_t = \argmax_{p \in [a_{j_t-1},a_{j_t}] } p \gamma \sqrt{ \phi_j(p )^{\top}  (\Lambda^j_{t,s})^{-1} \phi_j(p ) }
         $
         \State Update $ \Psi_{t+1}^s = \Psi_{t}^s \cup \{t\}$
         \EndIf
         \Until{ A price $p_t$ is found}
         \State Set the price $p_t $ and obtain the demand $d_t$
      \EndFor
   \end{algorithmic}  
\end{algorithm}

\textbf{Initialization.}
In each time period $t$, we begin by discrete the whole price range $[p_{min},p_{max}]$ into $N$ sub-intervals with equal length,
i.e., $[p_{min},p_{max}]=\bigcup_{j=1}^N[a_{j-1},a_j]$, where $a_j = p_{min} + j\frac{(p_{max} - p_{min})}{N}$ for $j=0,1,\cdots,N$.
We then can naturally construct the candidate action set $\mathcal{A}_{t,1} = [N]$. 
Such construction leads to linear bandit structure with bounded action set size.

\textbf{Local polynomial regression.}
For each $j\in [N]$, let $\theta^j $ be a $(\varpi (\beta)+1)$-dimensional column vector with its $k$-th element defined as $ \frac{f^{(k-1)}(a_j) }{(k-1)!} $.
According to \cref{lem: polynomial regression}, the demand function can be expressed as
$$
f(p) = \phi_j(p)^{\top}\theta^j+ \eta_j,
$$
where $\phi_j(p) = (1,p-a_j,\cdots,(p-a_j)^{\varpi (\beta)})^{\top}$ and $|\eta_j|\leq \frac{L(p_{max}-p_{min})^{\beta}}{N^{\beta}} $.
This formulation allows us to focus on the estimator for each $\theta^j$.

By using the local polynomial regression, we can leverage the H{\"o}lder smoothness condition to improve the approximation error at each small price interval. 
Suppose the length of a price interval is $\eta$. 
Under a constant approximation, the approximation error would be $\mathcal{O}(\eta)$ if the demand function $f$ is Lipschitz. 
Conversely, if $f$ satisfies \cref{asp: Smoothness}, 
we can bound the approximation error by $\mathcal{O}(\eta^{\beta})$ for $\beta>1$, 
thereby improving the original $\mathcal{O}(\eta)$ error. 

After the initialization step, each dataset $\mathcal{D}_{t,s}^j$ consists of at least $1+\varpi(\beta)$ samples.
Thanks to the polynomial map $\phi_j$, we can show that the first $1+\varpi(\beta)$ samples form a Vandermonde matrix,
which implies $\Lambda_{t,s}^j$ to be invertible.
\begin{myprop}
  \label{prop: invertible gram matrix}
  For each $t>N(1+\varpi(\beta))$, $s\in [S]$ and $j\in [N]$, the Gram matrix $\Lambda_{t,s}^j$ is invertible.
\end{myprop}

From \cref{prop: invertible gram matrix}, we know that the estimator $\theta_{t,s}^j$ is unique.
Based on the given estimators, we can construct the upper confidence bound for the optimal prices in each interval.
Some studies \citep{improvedUCB, linContextual, PlinearDP} concerning linear contextual bandits consider the following form: 
$$
\Lambda^j_{t,s} =  I_{1+\varpi(\beta)} +\sum_{\tau \in \mathcal{D}_{t,s}^j }  \phi_j(p_{\tau}) \phi_j(p_{\tau})^{\top}.
$$
However, this form is not applicable in our context, 
as it would cause the estimation error to depend on the upper bound of the linear parameter norm. 
In our scenario, it necessitates knowledge of the upper bounds of the derivatives of $f$, 
which can be unrealistic in practical applications.

\textbf{Layered Data Partitioning.}
During the round $t$, we utilize the local polynomial regression estimator $\theta_{t,s}^j$ to drive a UCB procedure that strikes a balance between $f$-learning and revenue maximization. 
To explore potential optimal prices $p \in [p_{min},p_{max}]$ efficiently, 
we focus our attention on the $f$-values within the range of $[p_{min},p_{max}]$. 

Instead of utilizing all the data collected prior to time $t$ for an ordinary least squares estimator for $\theta^j$, 
we divide the preceding time periods into disjoint layers indexed by $s\in [S]$, such that $\cup_{s \in [S]} \Psi_t^s = [t-1]$.
At time $t$, we sequentially visit the layers $s\in [S]$ and, 
within each layer, calculate the upper confidence bound for each $\theta^j$ using data collected exclusively during the periods in $\Psi_t^s$.

The algorithm commences with $s=1$, corresponding to the widest confidence bands, and incrementally increases $s$, while simultaneously eliminating sub-optimal prices using OLS estimates from $\Psi_t^s$.
This procedure is meticulously designed to ensure that the ``stopping layer" $s_t$ determined at time $t$ solely depends on $\{\Psi_t^s\}_{s\leq s_t}$, 
effectively decoupling the statistical correlation in OLS estimates. 
The validity of this property has been formally stated and proven in \citep{minimaxOptimalLin,supLinUCB}.

The UCB property described in \cref{lem: UCB} is highly intuitive. 
The first term captures the estimation, 
while the second term represents the bias stemming from the approximation error. 
As the learning progresses, the variance gradually diminishes. 
With an increasing number of samples of $p_t$ and $d_t$, our constructed $\theta_{t,s}^j$ approach to $\theta^j$ with the distance at most $\frac{L(p_{max}-p_{min})^{\beta} \sqrt{ (1+\varpi(\beta))\ln t} }{N^{\beta}}  $.
It is important to note that our constructed UCB does not involve the bias term. 
This allows us to eliminate the need for knowledge of the Lipschitz constant. 
The reason behind this is that we compare the UCB values on both sides, 
and the largest bias terms are the same for all indices.
Therefore, we can deduct the bias term at both sides in the step (a) of the algorithm and the Lipschitz term is implicitly absorbed in the adaptive term $p_{max}2^{1-s}$.
This advantage of our algorithm eliminates the need for explicit knowledge of the Lipschitz constant.

\begin{mylem}
    \label{lem: UCB}
    Consider the round $t$ and any layer $s\in [s_t]$ in the round $t$.
    Then for each $j\in[N]$, we have 
    $$
    |p(\theta^j -\theta^j_{t,s})^{\top} \phi_j(p) | \leq  \gamma p\sqrt{ \phi_j(p )^{\top}  (\Lambda^j_{t,s})^{-1} \phi_j(p ) }  + \frac{Lp_{max}(p_{max}-p_{min})^{\beta}  \sqrt{2 (1+\varpi(\beta))\ln t} }{ N^{\beta}  }   ,
    $$
    with probability at least $ 1- \frac{\delta }{NS T}$.
\end{mylem}

The proof of \cref{lem: UCB} is also intuitive.
Using polynomial regression, we implicitly construct a linear demand model given by $\tilde{d}=\phi_j(p)^\top\theta^j + \epsilon$.
In each round, the linear demand model and the true model share the same price and noise, but they differ only in the demand.
We then use the least squares estimator $\tilde{\theta}^j_{t,s}$ as a bridge to establish an upper bound between $\theta^j$ and $\theta^j_{t,s}$.
Since $\tilde{\theta}^j_{t,s}$ is an unbiased estimator of $\theta^j$,
the first term in the RHS of \cref{lem: UCB} arises from the estimation error. 
Because $\tilde{\theta}^j_{t,s}$ shares the same structure as ${\theta}^j_{t,s}$,
we can apply \cref{lem: polynomial regression} to bound the distance between these two estimators.
Combining the two bounds, we obtain the results mentioned above.

Based on \cref{lem: UCB}, we construct a high-probability event 
\begin{align*}
  \Gamma = & \Bigg\{ |p(\theta^j -\theta^j_{t,s})^{\top} \phi_j(p) | \leq  \gamma p\sqrt{ \phi_j(p )^{\top}  (\Lambda^j_{t,s})^{-1} \phi_j(p ) } \\ 
   & \quad\quad\quad + \frac{Lp_{max}(p_{max}-p_{min})^{\beta}  \sqrt{2 (1+\varpi(\beta))\ln t} }{ N^{\beta}  }, \forall t, \forall s,\forall j  \Bigg\}.
\end{align*}
Then by the union bound, we have 
$$
\prob{\overline{\Gamma}} \leq \frac{\delta  }{  NST } \times N \times T \times S = \delta .
$$

\subsection{Regret analysis}
In this subsection, we analyze the regret of our proposed \cref{alg: PADP}. 
We will show how to configure the hyperparameter $N$.

For simplicity, we introduce the revenue function
$$
rev(p) = pf(p).
$$
We define the index of the best price within the interval $[a_{j-1},a_j] $ at the candidate set $\mathcal{A}_{t,s}$ as 
$$
j_{t,s}^* = \argmax_{j\in \mathcal{A}_{t,s}} \sup_{p\in [a_{j-1},a_j] }  rev(p) .
$$
We also denote one of the best price as $p^* \in \argmax_{p\in [p_{min},p_{max}] }  rev(p) $.

One special case is the polynomial demand function.
In this case, the approximation error is exactly zero and 
our algorithm ensures $ rev( p^*   ) = rev( p_{t,s}^{j^*_{t,s}})   $ for each $s$,
as the optimal action will not be eliminated when their constructed high-probability event holds.
For general cases, due to the approximation error being proportional to $  \frac{L(p_{max}-p_{min})^{\beta}  \sqrt{2 (1+\varpi(\beta))\ln t} }{ N^{\beta}  }     $ when transitioning to a new layer,
the regret of setting the price indexed by $j_{t,s}^*$ is naturally proportional to $s-1$.
Combining the above discussion, we proceed to prove the following lemma. 

\begin{mylem}
  \label{lem: difference of best prices for each layer}
  Given the event $\Gamma$,
  then for each round $t$ and each layer $s\in [s_t-1]$, 
  we have 
  $$
   rev( p^*   ) - rev( p_{t,s}^{j^*_{t,s}})   \leq   \frac{2Lp_{max}(p_{max}-p_{min})^{\beta}  \sqrt{2 (1+\varpi(\beta))\ln t} }{ N^{\beta}  }    (s-1).
  $$
\end{mylem}

Now, we can effectively control the regret within layer $s$ using $j^*_{t,s}$ as the benchmark. 
The upper bound in \cref{lem: regret of prices at layer s} can be divided into two components: the variance and the bias. 
As we increase the value of $s$, the variance decreases exponentially, 
while the bias only increases linearly. 
Therefore, having a larger value of $s$ is advantageous for online learning.

\begin{mylem}
  \label{lem: regret of prices at layer s}
  Given the event $\Gamma$,
  then for each round $t$, 
  we have 
    \begin{align*}
      p^{j^*_{t,s}}_{t,s} \phi_{ j^*_{t,s}  } ( p^{j^*_{t,s}}_{t,s} )^{\top}\theta^{j^*_{t,s}}   -   p_t \phi_{j_t}(p_t)^{\top} \theta^{ j_t  }   \leq  6 p_{max} \cdot 2^{-s} +   \frac{Lp_{max}(p_{max}-p_{min})^{\beta}  \sqrt{2 (1+\varpi(\beta))\ln t} }{ N^{\beta}  } ,
    \end{align*}
  for all $2\leq s < s_t$.
\end{mylem}

Based on \cref{lem: difference of best prices for each layer} and \cref{lem: regret of prices at layer s},
it is straightforward to derive the upper bound of the discrete-part regrets for prices in the layer $s$.

\begin{mylem}
  \label{lem: discrete regret at the round t}
  Given the event $\Gamma$, for every round $t$, we have 
  $$
  rev( p^*   ) - rev( p_t ) \leq 12 p_{max} \cdot  2^{-s_t} + \frac{Lp_{max}(p_{max}-p_{min})^{\beta}  \sqrt{2 (1+\varpi(\beta))\ln t} }{ N^{\beta}  }    (2s_t-1).
  $$
\end{mylem}

Therefore, the remaining tasks is to count the rounds in each $\mathcal{D}_{t,s}^j$.
From \cref{lem: size of dataset j}, we know 
$$
|\mathcal{D}_{t,s}^j | \leq 2^s \gamma \sqrt{2 (1+\varpi(\beta)) | \mathcal{D}_{t,s}^j| \ln | \mathcal{D}_{t,s}^j|  },
$$
which implies 
$$
|\Psi_{T+1}^s| = \sum_{j=1}^N |\mathcal{D}_{T+1,s}^j| \leq \gamma^2 2^{2s+1} N (\varpi(\beta) +1) \ln T.
$$
Based on the previous given materials, we are ready to prove the following theorem.
\begin{mythm}
    \label{thm: regret upper bound}
    Under \cref{asp: Smoothness}, the expected regret of \cref{alg: PADP} with $N = \lceil  (p_{max}-p_{min})^{\frac{2\beta}{2\beta+1}}  T^{\frac{1}{2\beta+1}}     \rceil$ satisfies 
    $$
    \mathbb{E}[Reg(T)] = \mathcal{O} ( (1+  L  )p_{max} (p_{max}-p_{min})^{\frac{\beta}{2\beta+1}}  (\varpi(\beta)+1)^{\frac{1}{2}}  \log^{\frac{3}{2}}_2 (T/\delta ) T^{ \frac{\beta+1}{2\beta+1} } )   .
    $$
    with probability at least $1-\delta$.
\end{mythm}

\textbf{Lower bound.}
At a higher level, Theorem 2 in \citep{Multimodal_DP} shows that for a class of nontrivial problem instances, 
the optimal cumulative regret any admissible dynamic pricing policy could achieve is lower bounded by $\Omega(T^{ \frac{\beta+1}{2\beta+1} } )$ if the underlying demand function $f$ is assumed to satisfy \cref{asp: Smoothness}. 
This result shows that the regret upper bounds established in \cref{thm: regret upper bound} is optimal in terms of dependency on $T$ (up to logarithmic factors) and therefore cannot be improved.

We now discuss some special cases of $\beta$.

\textbf{Very Smooth Demand.}
An interesting special case of \cref{thm: regret upper bound} is when $\beta = +\infty$, 
implying that the underlying demand function $f$ is very smooth. 
In such a special case, it is recommended to select $\beta$ as $\beta = \ln T$,
and the regret upper bound in \cref{thm: regret upper bound} could be simplified to
$$
\mathcal{O}(p_{max} \sqrt{(p_{max}-p_{min})T}   \log^{\frac{3}{2}}_2 (T/\delta ) ) .
$$

\textbf{Strongly concave rewards.}
In various academic studies, researchers have examined the implications of strongly concave rewards. 
This concept implies that the demand function adheres to specific smoothness conditions, notably characterized by \cref{asp: Smoothness} with  $\beta=2$.
Moreover, the second derivatives of the reward function $pf(p)$ are negative.

The imposition of strong concavity on the reward function has been a prevalent assumption in previous works, 
such as those by \citep{PricingMultiUserItem} and \citep{PricingInventory}. 
This condition results in a uni-modal revenue function, which significantly simplifies the learning process.
In contrast, our method is specifically designed to handle multi-modal demand functions, which are more representative of real-world applications.

From these observations, we can establish the following assumption, which has been discussed in numerous papers, including \citep{Multimodal_DP} and \citep{PricingMultiUserItem}. We will present this assumption directly to avoid unnecessary repetition in our discussion.
\begin{myasp}
  \label{asp: concavity}
  The revenue function $rev(p)=pf(p)$ satisfies 
  $$
  c_1(p-p^*)^2 \leq rev(p^*) - rev(p) \leq c_2(p-p^*)^2
  $$
  for some positive constants $c_1$ and $c_2$.
\end{myasp}

To achieve the optimal regret order,
we can implement \cref{alg: PADP} with $\beta=2$ and $N = \Theta(T^{\frac{1}{4}})$.
By modifying the proof of \cref{lem: discrete regret at the round t}, we can establish that
$$
rev(p^*) - rev(p) \leq \mathcal{O}( \frac{s\sqrt{\log T}}{N^2}  )
$$
for any $p\in [a_{j-1},a_j]$ with $j \in \mathcal{A}_{t,s}$.
Next, we substitute the endpoints into \cref{asp: concavity} to derive that
$$
rev(p^*) - rev(p^{j_{t,s}}_{t,s}) \geq c_1( \frac{|j_{t,s}-j^*|-1}{N} )^2
$$
for $j_{t,s}\in\mathcal{A}_{t,s}-\{ j_{t,s}^* \}$.
It implies that $|j_{t,s}-j^*| \leq \mathcal{O}(\log\log T) $,
which further suggests that $\mathcal{A}_{t,s} =\mathcal{O}(\log\log T)  $. 

Therefore, during the proof of \cref{thm: regret upper bound}, 
we have 
$$
|\Psi_{t +1}^s| \leq \sum_{j\in \mathcal{A}_{t,s}} |\mathcal{D}_{t,s}^j| \leq \mathcal{O}(   2^s \gamma \sqrt{2 (\varpi(\beta)+1 )  | \Psi_{T+1}^s| \ln T \log\log T}),
$$
due to \cref{lem: size of dataset j}.
The regret due to approximation error can be bounded by
$
\mathcal{O}(T/N^2) = \mathcal{O}(\sqrt{T}).
$
By combining these findings, we conclude that our method yields
$
\tilde{\mathcal{O}}(\sqrt{T})
$
regret.

From the preceding discussion, we recognize that \cref{asp: concavity} is a strong condition that significantly reduces regret. Demand functions that satisfy this assumption exhibit performance characteristics similar to those of very smooth functions.

A notable case discussed in \citep{PricingMultiUserItem} examines the demand function $1-F(p)$ with Bernoulli feedback.   
Their method incurs 
$
\tilde{\mathcal{O}}(\sqrt{T})
$
regret, which matches the order of our results. 
However, our approach demonstrates advantages by generalizing to accommodate a broader class of demand functions.

\textbf{Comparison to existing works.}
If $\beta \in \mathbb{N}$,  a comparison can be made between our findings and those of \cite{Multimodal_DP}:
$$
\mathcal{O}( \beta ln(\beta T) (1+L) (p_{max}-p_{min})^{\beta}  T^{ \frac{\beta+1}{2\beta+1} }  ).
$$
Firstly, this bound requires $p_{max}-p_{min}\leq 1$. 
Otherwise, the upper bound will exponentially increase when $\beta$ goes to infty,
i.e., the demand function is supper smooth.
Secondly, we have enhanced the order of $\beta$ from $\tilde{\mathcal{O}}(\beta(p_{\text{max}} - p_{\text{min}})^{\beta} T^{\frac{\beta+1}{2\beta+1}})$ to $\tilde{\mathcal{O}}(\sqrt{\beta}(p_{\text{max}} - p_{\text{min}})^{\frac{\beta}{2\beta+1}} T^{\frac{\beta+1}{2\beta+1}}  )$. 
This improvement also eliminates the constraint of $p_{\text{max}} - p_{\text{min}} \leq 1$.
Additionally, our method does not require the knowledge $L$, compared with \citep{Multimodal_DP}.

\section{Linear Contextual Effect}

In this section, we study the semi-parametric demand model with linear contextual effect,
i.e., $rev(p_t,x_t) = f(p_t) + \mu^\top x_t$.
In this setting, $x$ represents the context, and $\mu$ is a context-dependent parameter.
Such model is also investigated by many scholars such as \citet{PlinearDP,DP_separable}.
We now introduce some assumptions on this model. 

\begin{myasp}
   \label{asp: regularity}
   \begin{enumerate}
      \item The context $x_t$ is drawn i.i.d. from some unknown distribution with support in the $\alpha$-dimensional unit ball.
      \item The Gram matrix $\Sigma = \mathbb{E}[xx^\top]$ is invertible, i.e., $\lambda_{min}(\Sigma)>0$.
   \end{enumerate}
   
\end{myasp}

The first assumption is standard in the dynamic pricing literature, see, e.g., \citep{PersonalizedDP,DP_Elasticity}.
The unit ball can be generalized into any compact sets.
The second assumption can be found in linear demand models \citep{PlinearDP,PersonalizedDP}
and it also applies to linear bandit setting \citep{GLbandit,improvedGB,lasso_bandit}.

To incorporate contextual information into our framework, we introduce an extended feature map:
$$
\varphi(p,x) = ( \phi(p)^{\top}, x^{\top} )^{\top},
$$
where $\phi(p)$ is the original polynomial map for the price $p$. 
The remaining steps in our approach are largely unchanged. 

When context is not present, the optimal price remains fixed.
In the absence of context, \citet{Multimodal_DP} propose the idea of embedding the dynamic pricing into a multi-armed bandit framework, treating each price segment as a distinct arm.
The segment to which the fixed optimal price belongs is regarded as the ``best" arm. 
In contrast, in our context-based model, the optimal price varies depending on the context revealed in each period and thus changes over time.
As a result, the ``best" arm shifts with each period, making the static multi-armed bandit approach described by \cref{alg: PADP} inapplicable in the new setting.
To address this challenge, we modify \cref{alg: PADP} and present the new algorithm in \cref{alg: CDP-LDP}.

\begin{algorithm}[htp]
   \algnewcommand{\LeftComment}[1]{\Statex  \(//\) \textit{#1}}
   \caption{Parameter Adaptive Contextual Dynamic Pricing}
   \label{alg: CDP-LDP}
   \begin{algorithmic}[1]
       \Require the discretization number $N$, the length $T$, the smoothness parameter $\beta$, the confidence parameter $\delta$ and the price bound $[p_{min}, p_{max}]$
       \State Let $a_j = p_{min} + j\frac{(p_{max} - p_{min})}{N}$ for $j=0,1,\cdots,N$
       \State Let $\varphi_j(p,x) = (1,p-a_j,\cdots,(p-a_j)^{\varpi (\beta)},x^{\top})^{\top} \in \mathbb{R}^{\alpha +\varpi (\beta)+1} $
       \State Set $S = \lceil \log_2 \sqrt{ T} \rceil$, $\gamma = \sqrt{ \frac{1}{2}\ln(2NST/\delta) }$, $\Psi_{t}^s = \emptyset$ for $s\in [S]$, $\mathcal{T}=0$
       \For{$j=1,\cdots,N$}
           \State Let $\Lambda^j = O$ and $\mathcal{D}^j=\emptyset$
           \While{$\Lambda^j$ is not invertible and $\mathcal{T} < T$}
               \State Observe the context $x_{\mathcal{T}} \in \mathbb{R}^{\alpha}$
               \State Set a distinct price $p_{\mathcal{T}}$ in $[a_{j-1},a_j]$ and obtain the demand $d_{\mathcal{T}}$
               \State Update the round collection $\Psi_{{\mathcal{T}}+1}^{\sigma} = \Psi_{\mathcal{T}}^{\sigma} \cup \{{\mathcal{T}}\}$ for all $\sigma\in [S]$
               \State Update the dataset $\mathcal{D}^j=\mathcal{D}^j\cup\{\mathcal{T}\}$
               \State Compute the Gram matrix         
               \[
               \Lambda^j =  \sum_{\tau \in \mathcal{D}^{j}}  \varphi_j(p_{\tau},x_{\tau}) \varphi_j(p_{\tau},x_{\tau})^{\top}
               \]
               \State Update the round collection $\Psi_{t+1}^{\sigma} = \Psi_{t}^{\sigma}$ for all $\sigma\in [S]$
               \State $\mathcal{T} = \mathcal{T} + 1$
           \EndWhile
       \EndFor
       \LeftComment{(continued)}
   \end{algorithmic}  
\end{algorithm}

\begin{algorithm}[htp]
   \algnewcommand{\LeftComment}[1]{\Statex \quad\quad  \(//\) \textit{#1}}
   \caption{Parameter Adaptive Contextual Dynamic Pricing (continued)}
   \label{alg:CDP-LDP-2}

   \begin{algorithmic}[1]
      \LeftComment{Continued from \cref{alg: CDP-LDP}}
       \For{ round $t = \mathcal{T}+1,2,\cdots,T$ }
           \State Observe the context $x_t \in \mathbb{R}^{\alpha}$ 
           \State Let $s=1$ and $\mathcal{A}_{t,1} = \{j\in[N]| p_{\tau}\in [a_{j-1},a_j], \tau \in [t-1] \}$
           \Repeat 
               \State Compute $\mathcal{D}_{t,s}^j = \{ \tau \in \Psi_t^s | p_{\tau}\in [a_{j-1},a_j] \}$ for all $j\in \mathcal{A}_{t,s}$
               \State For all $j\in\mathcal{A}_{t,s}$, compute the estimator 
               \[
               \theta^j_{t,s} =  \argmin_{\theta }  \sum_{\tau \in \mathcal{D}_{t,s}^j }{ (d_{\tau} - \theta^\top \varphi_j(p_{\tau},x_{\tau})   )^2 }
               \]
               \State For all $j\in\mathcal{A}_{t,s}$, compute the matrix
               \[
               \Lambda^j_{t,s} =  \sum_{\tau \in \mathcal{D}_{t,s}^j }  \varphi_j(p_{\tau},x_{\tau}) \varphi_j(p_{\tau},x_{\tau})^{\top}
               \]
               \State Compute the upper confidence bound 
               \[
               U^j_{t,s}(p) =  \varphi_j(p,x_{t})^{\top} \theta^j_{t,s} + \gamma \sqrt{ \varphi_j(p,x_{t})^{\top}  (\Lambda^j_{t,s})^{-1} \varphi_j(p,x_{t}) }
               \]
               \If{ $ \sup_{ p \in [a_{j-1},a_j] } p \sqrt{ \varphi_j(p,x_{t})^{\top}  (\Lambda^j_{t,s})^{-1} \varphi_j(p,x_{t}) } \leq p_{max}/\sqrt{T} $ for all $ j \in \mathcal{A}_{t,s} $}
                   \State Choose the price
                   $
                   p_t = \argmax_{j\in [N]}\sup_{p \in [a_{j-1},a_j]} p U^j_{t,s}(p)
                   $
                   \State Update the round collection $\Psi_{t+1}^{\sigma} = \Psi_{t}^{\sigma}$ for all $\sigma\in [S]$
                   \LeftComment{step (a)}
               \ElsIf{ $ \sup_{p\in [a_{j-1},a_j]}  p \sqrt{ \varphi_j(p,x_{t})^{\top}  (\Lambda^j_{t,s})^{-1} \varphi_j(p,x_{t}) } \leq p_{max} 2^{-s} $ for all $j \in \mathcal{A}_{t,s}$ }
                   \State Let $\mathcal{A}_{t,s+1} = \{ j \in \mathcal{A}_{t,s}| \sup_{ p \in [a_{j-1},a_j]} p  U^j_{t,s}(p)  \geq  \max_{  j' \in \mathcal{A}_{t,s}} \sup_{p\in [a_{j-1},a_j]} p U^{j'}_{t,s}(p)  -  p_{max} 2^{1-s} \}$
                   \State Let $s \leftarrow s+1$
                   \LeftComment{step (b)}
               \Else
                   \State Choose $j_t$ such that $\sup_{p\in [a_{j_{t}-1},a_{j_t}]} p \gamma \sqrt{ \varphi_j(p,x_{t})^{\top}  (\Lambda^j_{t,s})^{-1} \varphi_j(p,x_{t}) } >   p_{max} 2^{-s}$
                   \State Choose the price 
                   $
                   p_t = \argmax_{p \in [a_{j_t-1},a_{j_t}] } p \gamma \sqrt{ \varphi_j(p,x_{t})^{\top}  (\Lambda^j_{t,s})^{-1} \varphi_j(p,x_{t}) }
                   $
                   \State Update $ \Psi_{t+1}^s = \Psi_{t}^s \cup \{t\}$
               \EndIf
           \Until{ A price $p_t$ is found}
           \State Set the price $p_t$ and obtain the demand $d_t$
       \EndFor
   \end{algorithmic}
\end{algorithm}

Unlike the initialization step in \cref{alg: PADP}, we use a stopping condition to guarantee the invertiblity of required Gram matrices.
The following lemma indicates the expected stopping time $\mathcal{T}$ is of order $\mathcal{O}(N\log T)$.
\begin{mylem}
   \label{lem: stopping time}
   In \cref{alg: CDP-LDP}, the following holds 
   $$
   \mathbb{E}[\mathcal{T}] \leq C_0N\log T,
   $$
   for some positive constant $C_0$.
   The constant $C_0$ depends on $\alpha,\beta$ and $\lambda_{min}(\Sigma)$.
\end{mylem}

Similar to \cref{thm: regret upper bound}, we can bound the regret of \cref{alg: CDP-LDP},
which is presented in \cref{thm: regret upper bound of contextual dynamic pricing algorithm}.
\begin{mythm}
      \label{thm: regret upper bound of contextual dynamic pricing algorithm}
      Under \cref{asp: Smoothness} and \cref{asp: regularity}, the expected regret of \cref{alg: CDP-LDP} with $N = \lceil  (p_{max}-p_{min})^{\frac{2\beta}{2\beta+1}}   T^{\frac{1}{2\beta+1}}     \rceil$ satisfies 
      $$
      \mathbb{E}[Reg(T)] = \mathcal{O} ( (1+  L  )p_{max} (p_{max}-p_{min})^{\frac{\beta}{2\beta+1}}  (\alpha+\varpi(\beta)+1)^{\frac{1}{2}}   \log^{\frac{3}{2}}_2 (T/\delta ) T^{ \frac{\beta+1}{2\beta+1} } )   .
      $$
      with probability at least $1-\delta$.
\end{mythm}

The proof of \cref{thm: regret upper bound of contextual dynamic pricing algorithm}
 is almost identical to that of \cref{thm: regret upper bound}.
 The difference is to replace the feature map $\phi$ with the new one $\varphi$.
 Another difference lis in the initialization step.
 Since the contexts are i.i.d. generated, we use the stopping time analysis to guarantee the invertiblity.
 For completeness, we defer the proofs of \cref{thm: regret upper bound of contextual dynamic pricing algorithm} in Appendix.

 \textbf{Comparison to existing works.}
 This improvement also shows advantageous in the linear contextual effect setting.
 As shown in \cite{PlinearDP}, the regret is upper bounded by
 $$
 \tilde{\mathcal{O}}( (\alpha+\varpi(\beta)+1) L^{  \frac{1}{2\beta+1}  } T^{ \frac{\beta+1}{2\beta+1} }  ).
 $$
 We have elevated the order of $\alpha + \varpi(\beta) + 1$ to $\sqrt{\alpha + \varpi(\beta) + 1}$. 
 Regarding the parameter $L$, the methodology in \cite{PlinearDP} mandates prior knowledge of $L$. 
 By configuring the parameter:
 $$
 N =\lceil L^{\frac{2}{2\beta+1}} (p_{max}-p_{min})^{\frac{2\beta}{2\beta+1}}  ( T \log_2 T )^{\frac{1}{2\beta+1}}     \rceil 
 $$
 with knowledge of $L$, we can still attain the optimal order concerning $L$. 
 However, in cases where $L$ is unknown, our approach can adapt to this parameter, 
 in contrast to the method in \cite{PlinearDP}.
 Besides, the method in \citep{PlinearDP} is not rate-optimal in terms of the dimension $\alpha$.

 Additionally, in the linear demand model $d(p, x) = bp + \mu^{\top}x$, the upper bound transforms to $\tilde{\mathcal{O}}(\sqrt{\alpha T})$, 
 outperforming the optimal regret rate $\tilde{\mathcal{O}}(\alpha \sqrt{T})$ in \citep{PersonalizedDP} in terms of the context dimension $\alpha$.
 \citet{DP_Elasticity} proves that the lower bound of this problem is $\Omega(\sqrt{\alpha T})$,
 which indicates that our method can achieve the minimax optimality in this simple demand function.
 
\subsection{Adaptivity of smoothness parameter}
When the smoothness parameter $\beta$ is unknown, non-adaptive dynamic pricing algorithms cannot reach the optimal regret rate.
It has been shown in \citep{adaptivity_impossible} that no strategy can adapt optimally to the smoothness level of $f$ for cumulative regret.
However, adaptive strategies that are minimax optimal can exist if additional information about $f$ is available.
The self-similarity condition \citep{self_similarity,SmoothnessApaptiveDP,TLinCB,SmoothnessApaptiveCB} is often used to achieve adaptivity with beneficial properties.

To formalize this concept, we first define notation for function approximation over polynomial spaces.
For any positive integer $l$, $Poly (l)$ represents the set of all polynomials with degree up to $l$. 
For any function $g(\cdot)$, we use $\Gamma_{l}^Ug(\cdot)$ to denote its $L^2$-projection onto $Poly (l)$ over some interval $U$, 
which can be obtained by solving the minimization problem:
\begin{align*}
  \Gamma_{l}^Ug(x) = \min_{q\in Poly (l)} \int_U |g(u)-q(u)|^2 du.
\end{align*}

\begin{mydef}
   A function $g: [a, b] \to \mathbb{R} $ is self-similar on $[a, b]$ with parameters $\beta > 1$, $\ell \in \mathbb{Z}^+$, $M_1\geq 0$ and $M_2> 0$ 
   if for some positive integer $c > M_1$ it holds that
$$
\max_{V\in \mathcal{V}_c} \sup_{x \in V} |\Gamma_{\ell}^V g(x) -g(x)| \geq M_2 2^{-c\beta},
$$
   where we define
$$
\mathcal{V}_c = \left\{   [ a+\frac{i(b-a)}{2^c}, a+\frac{(i+1)(b-a)}{2^c} ] , i=0,1,\cdots,2^c-1  \right\}
$$
   for any positive integer $c$. 
\end{mydef}

Unlike H{\"o}lder smoothness, the self-similarity condition sets a global lower bound on approximation error using polynomial regression. 
This approach helps estimate the smoothness of demand functions by examining approximations across different scales.
It has been used in nonparametric regression to create adaptive confidence intervals.
In previous work, this condition was applied to estimate non-contextual demand functions \citep{SmoothnessApaptiveDP}, and for general contextual demand functions \citep{TLinCB}.
However, because our contextual terms are additive, we can reduce sample usage and achieve minimax optimality.

\begin{algorithm}[htp]
	\renewcommand{\algorithmicrequire}{\textbf{Input:}}
	\renewcommand{\algorithmicensure}{\textbf{Output:}}
    \algnewcommand{\LeftComment}[1]{\Statex \quad\quad  \(//\) \textit{#1}}
	\caption{Smoothness Parameter Estimation}
	\label{alg: Smoothness Parameter Estimation}
	\begin{algorithmic}[1]
      \Require the length $T$, the smoothness parameter upper bound $\beta_{max}$, the price bound $[p_{min},p_{max}]$
      \State Set local polynomial regression degree $\varpi(\beta_{max})$
      \State Set $k_1 = \frac{1}{2\beta_{max}+2}$, $k_2 = \frac{1}{4\beta_{max}+2}$, $K_1 = 2^{ \lfloor k_1\log_2 T  \rfloor   }$, $K_2 = 2^{ \lfloor k_2\log_2 T  \rfloor   }$
      \For{ $i = 1,2$ }
         \State Set trial time $T_i = T^{ \lfloor \frac{1}{2} + k_i  \rfloor   }$
         \State Set the price $T_i$ times from $U[p_{min},p_{max}]$ independently
         \State Collect samples $(x_j,p_j,d_j)$ with size $T_i$
         \For{ $m = 1,2,\cdots,K_i$ }
         \State Fit local polynomial regression on $[p_{min}+\frac{(m-1)(p_{max}-p_{min})}{K_i},p_{min}+\frac{m(p_{max}-p_{min})}{K_i}]$ with samples falling in the interval
         \State Construct the estimate $\hat{f}_i(p)$ on $[p_{min}+\frac{(m-1)(p_{max}-p_{min})}{K_i},p_{min}+\frac{m(p_{max}-p_{min})}{K_i}]$ and corresponding $\hat{\mu}_i$ from the regression
         \EndFor
      \EndFor
      \Ensure Let $\hat{\beta} = -\frac{\ln( \max_{p,x}{|\hat{f}_2+\hat{\mu}_2^\top x-\hat{f}_1-\hat{\mu}_1^\top x|}  )}{\ln T} - \frac{\ln \ln T}{\ln T} $
   \end{algorithmic}  
\end{algorithm}

Our algorithm is based on a clear concept. 
We start by dividing the price range into small intervals, with each interval's length determined by $\beta_{max}$. 
This enables us to use local polynomial regression to approximate the true demand function within each segment. 
As shown in \cref{col: adaptivity to smoothness level}, our non-adaptive algorithm reaches optimal regret when the correct smoothness parameter is applied.

The rationale behind \cref{alg: Smoothness Parameter Estimation} hinges on leveraging the H{\"o}lder smoothness property of the demand function, 
which justifies the use of local polynomial regression for estimation. 
Nevertheless, the self-similarity intrinsic to H{\"o}lder smoothness functions complicates polynomial-based approximations,
as it imposes recursive regularity constraints across scales. 
To mitigate this, the algorithm refines its estimation granularity—i.e., the number of subintervals partitioning the price domain—to enhance the resolution of the demand function's approximation. 
By iteratively adjusting the partition density, the method achieves a high-fidelity piecewise polynomial representation, 
balancing local accuracy with global structural constraints.

\begin{mylem}
   \label{lem: estimation upper bound}
   Let $\mathbb{O}_{[a,b]}=\{(x_i,p_i,d_i)\}_{i=1}^n$ be an i.i.d. sample set, where each $p_i \in [a,b]$ and $\|x_i\| \leq 1$.
   Assume the sub-Gaussian parameter $u_1 \leq \exp(u_1' n^v)$ for some positive constants $v$ and $u_1'$, and that the polynomial degree $\ell \geq \alpha + \varpi(\beta)$. 
   Then, there exist positive constants $C_1$ and $C_2$ such that, with probability at least $1 - \mathcal{O}(\ell e^{-C_2 \ln^2 n})$, for any $p \in [a,b]$, any $\norm{x}_2\leq 1$ and $n > C_1$, the following inequality holds:
   $$
   |f(p)+\mu^\top x - \hat{f}(p) - \hat{\mu}^\top x| < (b-a)^{\beta}\ln n + n^{-\frac{1}{2}(1-v)} \ln^3 n,
   $$
   where $\hat{f}$ and $\hat{\mu}$ are estimated from the sample $\mathbb{O}_{[a,b]}$.
\end{mylem}

\Cref{lem: estimation upper bound} shows that the estimation error in the linear contextual effect setting. 
If we apply the method from \citep{TLinCB}, the approximation error is $\mathcal{O}((b-a)^{\alpha+\beta})$. 
However, due to the additive structure of the demand function, we can avoid the extra $\alpha$ term in the error bound.

\begin{mythm}
   \label{thm: beta confidence bound}
   With an upper bound $\beta_{max}$ of the smoothness parameter, under the assumptions in our setting,
   for some constant $C>0$, with probability at least $1-\mathcal{O}(e^{-C\ln^2 T})$, 
   $$
   \hat{\beta} \in [\beta- \frac{4(\beta_{max}+1)\ln \ln T}{\ln T}   ,\beta].
   $$
\end{mythm}

\Cref{thm: beta confidence bound} highlights the effectiveness of our proposed \cref{alg: Smoothness Parameter Estimation} in estimating the H{\"o}lder smoothness parameter $\beta$.
The adaptability and efficient smoothness parameter estimation enable it to construct tight confidence intervals for $\beta$ and achieve a high convergence rate. 
These features contribute to the desired regret bound in dynamic pricing scenarios and open avenues for developing more robust and adaptive pricing algorithms.

The estimators of the demand function, denoted $\hat{f}_1$ and $\hat{f}_2$, allow us to determine an estimate for the H{\"o}lder smoothness parameter $\hat{\beta}$.
The estimator $\hat{\beta} $ is then fed into \cref{alg: CDP-LDP} for the remaining time horizon $T-T_1-T_2$. 
We notice that for demand function $f$ satisfying \cref{asp: Smoothness} with the parameter $\beta$,
it also satisfies \cref{asp: Smoothness} with the parameter $\hat{\beta}\leq \beta$.
Therefore, if \cref{alg: PADP} is invoked with the estimator $\hat{\beta}\leq \beta$,
then the regret is upper bounded by 
$$
\mathcal{O} ( (1+  L  )p_{max} (p_{max}-p_{min})^{\frac{\hat{\beta}}{2\hat{\beta}+1}}    (\log_2 T )^{\frac{4\hat{\beta} +3}{4\hat{\beta} +2}} T^{ \frac{\hat{\beta}+1}{2\hat{\beta}+1} } )   .
$$
From \cref{lem: concentration of beta}, we know that $\hat{\beta}$ converges to $\beta$ with rate $\mathcal{O}(\frac{\ln\ln T}{\ln T}  )$.
We know that 
$$
 \frac{\hat{\beta}+1}{2\hat{\beta}+1} - \frac{ {\beta}+1}{2{\beta}+1} = \frac{2(\beta-\hat{\beta})}{(2\hat{\beta}+1)(2{\beta}+1)}\leq \frac{8(\beta_{max}+1)\ln \ln T}{\ln T}.
$$
Therefore, we know that the regret caused by estimating $\beta$ is at most 
$$
\tilde{\mathcal{O}}(T^{\frac{\beta+1}{2\beta+1}} (\ln T)^{ 8(\beta_{max}+1) }).
$$

For the adaptive part, note that $T_1,T_2 \leq T^{\frac{\beta+1}{2\beta+1}}$, 
which means that the regret is upper bounded by $\mathcal{O}(T^{\frac{\beta+1}{2\beta+1}})$. 
We combing the above discussion we obtain the following corollary.
\begin{mycol}
   \label{col: adaptivity to smoothness level}
If we run \cref{alg: Smoothness Parameter Estimation} to obtain an estimator $\hat{\beta}$ and feed it to \cref{alg: PADP}, then the regret is 
$$
\mathbb{E}[Reg(T)] = \mathcal{O} \left( (1+  L  )p_{max} (p_{max}-p_{min})^{\frac{\beta}{2\beta+1}}    \log_2^{\frac{3}{2}} (T/\delta)  T^{ \frac{\beta+1}{2\beta+1} }  (\ln T)^{ 8(\beta_{max}+1) } \right) .
$$
\end{mycol}
Compared with \cref{thm: regret upper bound}, we pay extra $(\ln T)^{ 8(\beta_{max}+1) } $ term to obtain the estimator of $\beta$.

\subsection{Application to pricing competitions}
We now demonstrate how our method can be applied to the pricing competition models proposed in \citep{LEGO,concave_pc}.
These models consider a market with $K$ sellers, 
each offering a single product with unlimited inventory, over a selling horizon of $T$ rounds. 
At the beginning of each round, every seller simultaneously sets their price and then observes their private demand, 
which is influenced by the prices set by all other sellers.
The demand follows a noisy and unknown linear model. Specifically, the demand faced by seller $i$ is given by:
\begin{align}
   \label{eq: demand for seller i}
   d_i = f_i(p_i) + \sum_{j\neq i} \mu_{ij} p_j + \epsilon_i,
\end{align}
where $d_i,p_i,\epsilon_i$ are the demand, price and sub-gaussian noise for the seller $i$, respectively.
The competitors parameters $\mu_{ij}$ capture the effect of the price set by competitor $j$ on the demand faced by seller $i$. 
In \citep{LEGO}, an exact solution is proposed for this linear demand scenario, with a policy that achieves a regret bound of $\mathcal{O}(K\sqrt{T}\log T)$ regret for each seller.

The pricing competition model above relies on several key assumptions that may not hold in real-world markets. 
First, it assumes that each seller can observe the prices of all other competitors, 
which is often unrealistic. A more reasonable assumption would be to allow each seller to observe only a small subset of the prices set by other sellers. 
Second, in highly competitive markets, the prices of many sellers tend to stabilize over time, 
forming an invariant distribution. 
Therefore, it is unrealistic to expect all competitors to engage in an exploration phase, 
especially when a new seller enters the market.
Finally, the regret of \cref{alg: CDP-LDP} for each seller scales linearly with the number of competitors, $K$, 
which becomes problematic when $K$ is large. 
In practice, markets often consist of a very large number of sellers, 
making this result less satisfactory.

To address the limitations of the previous model, we consider the asymptotic regime, 
where the number of sellers $K$ approaches infinity. 
In this setting, we aim to provide an approximate solution for cases where $K$ is sufficiently large. 
We make the following assumptions:
\begin{myasp}
   \begin{itemize}
      \item The market is highly competitive and the market price distribution follows $F$.
      \item A single seller's price changes do not affect the overall price distribution $F$.
   \end{itemize}
\end{myasp}

Now, consider the approximation of the demand function in equation \eqref{eq: demand for seller i}. This leads to the following simplified model:
$$
d = f(p) + \mu^\top \mathbf{p}_o + \epsilon,
$$
where $\mathbf{p}_o \in \mathbb{R}^{\alpha} $ is a vector of prices observed by the seller, each component sampled from the joint market price distribution $F$.
In this model, we allow the seller to observe $\alpha$ prices from the market, 
which can be interpreted as an $\alpha$-dimensional context.
Without loss of generality, we assume the components of $\mathbf{p}_o$ are sorted, 
as this allows us to focus on the main effect of competition — namely, the values of the prices themselves — while ignoring the finer details of individual seller interactions.
In this framework, we abstract away from the specific interactions between sellers and instead focus on the relationship between the seller and the overall market price distribution.

By making slight modifications to the model in \citep{LEGO}, we arrive at a refined approximation. 
This model can be solved using the approach outlined in \cref{alg: CDP-LDP}. 
In the original model from \cite{LEGO}, the demand function is linear in the proposed price,
allowing us to apply the technique for very smooth demands and achieve a regret bound of $\mathcal{O}(\sqrt{\alpha T}\log T)$ regret.
Importantly, this regret is independent of the number of sellers, $K$. 
However, our approach extends to a broader range of demand functions, 
making it more flexible and applicable to a wider variety of market scenarios.

\section{Discussion}
\label{sec: discussion}

In this section, we explore several extensions and improvements to the existing framework, 
aiming to enhance our understanding of the theoretical results and their applications.

\textbf{Many products and customers.}
We consider $K$ types of cases, such as different customer or product categories.
The seller observes $K$ distinct contexts to differentiate between these cases. 
Each case has its own demand function $f_i(p)$, 
and the seller's goal is to maximize the corresponding revenue $p f_i(p)$ when case $i$ occurs.
Let $T_i$ denote the number of rounds when the case $i$ occurs.
A straightforward maximization problem leads to the following inequality:
$$
\sum_{i=1}^K  T_i^{  \frac{\beta+1}{2\beta+1}} \leq  K^{  \frac{\beta}{2\beta+1}} T^{  \frac{\beta+1}{2\beta+1}}.
$$
Furthermore, the environment can equally divide the whole rounds into $K$ equal length.
Then from the lower bound of nonparametric dynamic pricing \cite{Multimodal_DP}, we deduce a lower bound for $K$ cases
$$
\Omega(K\times (\frac{T}{K})^{  \frac{\beta+1}{2\beta+1}}) = \Omega(K^{  \frac{\beta}{2\beta+1}} T^{  \frac{\beta+1}{2\beta+1}}),
$$
as the lower bound in the single case is $\Omega( T^{  \frac{\beta+1}{2\beta+1}} )$
Notably, under the assumption of concavity (as specified in \cref{asp: concavity}), we achieve a regret bound of $\tilde{\mathcal{O}}(\sqrt{KT})$. 
In \cite{PricingMultiUserItem}, the authors utilize the demand function $1 - F_i(p)$ for each product-customer pair. 
It is important to note that in \cite{PricingMultiUserItem}, the total rounds can be viewed as $LT$, where $L$ denotes the maximum load parameter in their analysis.
If we let $K$ represent the number of such product-customer pairs, we can recover the same order of regret as shown in their work.

\textbf{Model misspecification.}
Our algorithms are robust to model misspecification. 
Suppose that the noise term has non-zero mean, and $|\epsilon_t|\leq \varepsilon$.
If we apply the methods in \cite{PlinearDP,Multimodal_DP},
we will require the knowledge of $\varepsilon$.
However, our method is parameter-free, meaning that we do not require $\varepsilon$.

\begin{mylem}
   \label{lem: UCB with model misspecification}
   Consider the round $t$ and any layer $s\in [s_t]$ in the round $t$.
   Then for each $j\in[N]$, we have 
   \begin{align*}
      |p(\theta^j -\theta^j_{t,s})^{\top} \phi_j(p) | \leq & \gamma p\sqrt{ \phi_j(p )^{\top}  (\Lambda^j_{t,s})^{-1} \phi_j(p ) } \\ 
      & + \frac{Lp_{max}(p_{max}-p_{min})^{\beta}  \sqrt{2 (1+\varpi(\beta))\ln t} }{ N^{\beta}  } + \varepsilon T p_{max}  \sqrt{ 2 (1+\varpi(\beta))\ln t  }  ,
   \end{align*}
   with probability at least $ 1- \frac{\delta }{NS T}$.
\end{mylem}

The proof steps are similar to the well-specified cases,
so we directly write down the following corollary.
If the misspecification error is small, i.e., $\varepsilon = \mathcal{O}(T^{-\frac{\beta}{2\beta+1}} )$,
thus our method can still achieve the minimax optimality,
indicating the robustness of our algorithm.

\begin{mycol}
   If the demand function is misspecified and $|\epsilon_t|\leq \varepsilon$ for all $t$, 
   then the regret of \cref{alg: PADP} is upper bounded by 
   $$
  \mathcal{O} ( (1+  L  )p_{max} (p_{max}-p_{min})^{\frac{\beta}{2\beta+1}}  (\alpha+\varpi(\beta)+1)^{\frac{1}{2}}  (\log_2 T )^{\frac{4\beta +3}{4\beta +2}} T^{ \frac{\beta+1}{2\beta+1} } + \epsilon T \sqrt{ (\alpha+\varpi(\beta)+1) \log T } )   
   $$
   with probability at least $1-\delta$.
\end{mycol}

\textbf{Unknown time horizon.}
We show that our algorithms can adapt to unknown time horizon $T$ by doubling trick.
Let the algorithms proceed in epochs with exponentially increasing in rounds. 
For $n$-th epoch, we run our algorithms with rounds $2^n$ and configure corresponding hyperparameter $N$ with such epoch length.
Then we can derive the following upper bound for \cref{alg: PADP}.
From Holder's inequality, we have
$$
\sum_{n=1}^{ \lceil \log_2 T \rceil } T_n^{\frac{\beta+1}{2\beta+1}} \leq T^{\frac{\beta+1}{2\beta+1}} (\log T)^{\frac{\beta}{2\beta+1}}.
$$
Therefore, we only need to pay extra $ (\log T)^{\frac{\beta}{2\beta+1}}$ terms to make our algorithms to adapt to unknown horizon $T$.

\section{Numerical Experiments}

In this section, we conduct numerical experiments to study the empirical performance of \cref{alg: PADP}. 
We measure the performance of a learning algorithm by the simple regret defined as
$
\frac{Reg(T)}{T}.
$
A similar metric is also utilized in \cite{Multimodal_DP} and \cite{PlinearDP}.

In the first numerical experiment, we consider the following demand function 
$$
f(p) = 1- \frac{1}{\beta}(\frac{p-p_{min}}{p_{max}-p_{min}})^{\beta}
$$
with added noise $\mathcal{N}(0,0.01)$.
We set the price range $[p_{min},p_{max}]=[2.6,3.8]$ and choose $\beta=2.5$. 
The tight Lipschitz constant for this setting is $L^*=\max_{p\in[p_{min},p_{max}]}|f'(p)|\approx 2.66$.

To demonstrate the advantages of our method, we compare it with the algorithm from \cite{Multimodal_DP},
tested with $L\in\{L^*,4,6,8,10\}$.
Both algorithms are evaluated over horizons 
$$T \in \{8000,10000,12000,14000,16000\}$$ 
for $50$ runs. 

\begin{figure}[htbp]
    \centering
    \begin{tikzpicture}
        \begin{customlegend}[legend columns=5,legend style={draw=none,column sep=2ex,nodes={scale=0.6, transform shape}},
            legend entries={
                            \text{\cite{Multimodal_DP}: $L=3$},
                            \text{\cite{Multimodal_DP}: $L=6$},
                            \text{\cite{Multimodal_DP}: $L=9$},
                            \text{\cref{alg: PADP}}
                            }]
    
            \addlegendimage{color=yellow}
            \addlegendimage{color=blue}
            \addlegendimage{color=green}
            \addlegendimage{color=purple}
            \addlegendimage{color=black}
        \end{customlegend}
    \end{tikzpicture}

    \begin{tabular}{lr}
        
    \centering

    \begin{tikzpicture}[scale=0.8]
        \begin{axis}[
            height = 0.4\textwidth,
            width = 0.6\textwidth,
            xlabel = time $t$,   
            ylabel = simple regret,
            xmin=8000,
            xtick pos = left,
            ytick pos = left,
            yticklabel style={/pgf/number format/.cd, fixed, fixed zerofill, precision=4},
        ]

        \addplot [smooth, mark=triangle, yellow ] table [x index=1,y index=2, col sep = comma] {data/poly_L=3.csv};
        \addplot [smooth, mark=o, blue ] table [x index=1,y index=2, col sep = comma] {data/poly_L=6.csv};
        \addplot [smooth, mark=square, green ] table [x index=1,y index=2, col sep = comma] {data/poly_L=9.csv};

        

        \end{axis}
    \end{tikzpicture}
    &
    \begin{tikzpicture}[scale=0.8]
        \begin{axis}[
            height = 0.4\textwidth,
            width = 0.6\textwidth,
            xlabel = time $t$,   
            xmin=8000,
            xtick pos = left,
            ytick pos = left,
        ]

        \addplot [smooth, mark=triangle, cyan] table [x index=1,y index=2, col sep = comma] {data/adaDP_poly_reg.csv};

        

        \end{axis}
    \end{tikzpicture}
\end{tabular}
 
\caption{Relative regret curve with H{\"o}lder smooth demand. }
\label{fig: relative regret curve}
\end{figure}
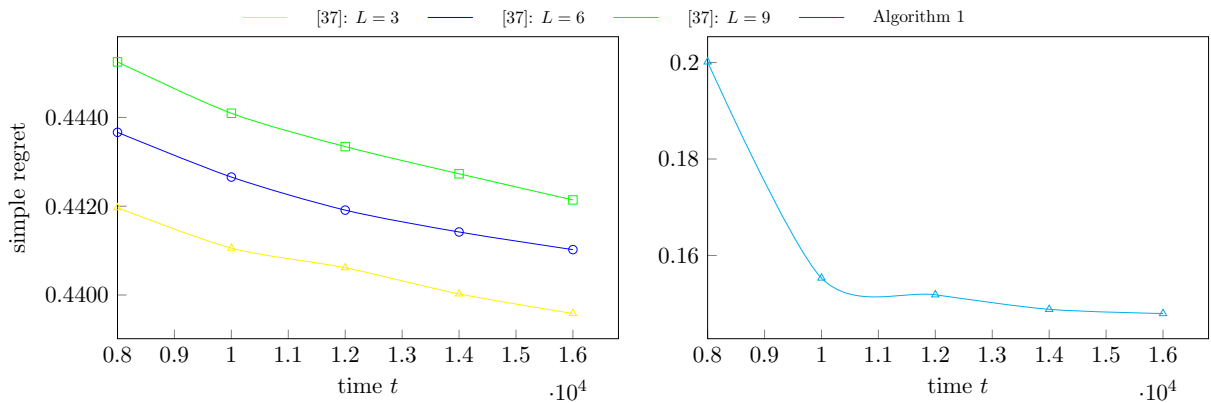

From \cref{fig: relative regret curve}, it is evident that larger Lipschitz constants lead to higher regret for the method in \cite{Multimodal_DP}. 
This underscores the importance of accurately knowing $L$, 
which can be unrealistic in real-world applications. 
In contrast, our method shows superior performance and practicality, 
as evidenced by its relative regret curve, outperforming the baseline algorithms in both effectiveness and ease of use.

In the second numerical experiment, we consider the following demand function
$$
1-0.5 \Phi((p-p_{min})/2)-0.5\Phi((p-p_{max})/2),
$$
where $\Phi$ denotes the cumulative distribution function of the standard normal distribution.
This demand function guarantees the unimodality of the revenue function.
Additionally, because this demand function is significantly smoother compared to the previous one, we set $\beta = \ln T$ in both algorithms.

We observe that the relative regrets of both algorithms decrease as the smoothness level increases.
Despite this, our method consistently outperforms the method in \cite{Multimodal_DP}, demonstrating superior performance even without prior knowledge of the smoothness parameter.

\begin{figure}[htbp]
    \centering
    \begin{tikzpicture}
        \begin{customlegend}[legend columns=6,legend style={draw=none,column sep=2ex,nodes={scale=0.6, transform shape}},
            legend entries={
                            \text{\cite{Multimodal_DP}: $L=1$},
                            $L=3$,
                            $L=5$,
                            $L=9$,
                            \text{\cref{alg: PADP}}
                            }]
    
            \addlegendimage{color=green}
            \addlegendimage{color=orange}
            \addlegendimage{color=blue}
            \addlegendimage{color=purple}
            \addlegendimage{color=cyan}
        \end{customlegend}
    \end{tikzpicture}

    \begin{tabular}{lr}
        
    \centering

    \begin{tikzpicture}[scale=0.8]
        \begin{axis}[
            height = 0.4\textwidth,
            width = 0.6\textwidth,
            xlabel = time $t$,   
            ylabel = simple regret,
            xmin=10000,
            xtick pos = left,
            ytick pos = left,
        ]

        \addplot [smooth, mark=diamond, green ] table [x index=1,y index=2, col sep = comma] {data/L=1.csv};
        \addplot [smooth, mark=square, orange ] table [x index=1,y index=2, col sep = comma] {data/L=3.csv};
        \addplot [smooth, mark=o, blue ] table [x index=1,y index=2, col sep = comma] {data/L=5.csv};
        \addplot [smooth, mark=star, purple ] table [x index=1,y index=2, col sep = comma] {data/L=9.csv};
               

        

        \end{axis}
    \end{tikzpicture}
    &
    \begin{tikzpicture}[scale=0.8]
        \begin{axis}[
            height = 0.4\textwidth,
            width = 0.6\textwidth,
            xlabel = time $t$,   
            xmin=10000,
            xtick pos = left,
            ytick pos = left,
        ]

        \addplot [smooth, mark=triangle, cyan ] table [x index=1,y index=2, col sep = comma] {data/adaDP_reg_ratio.csv};

        

        \end{axis}
    \end{tikzpicture}
\end{tabular}
 
\caption{Relative regret curve with the very smooth demand.}

\end{figure}
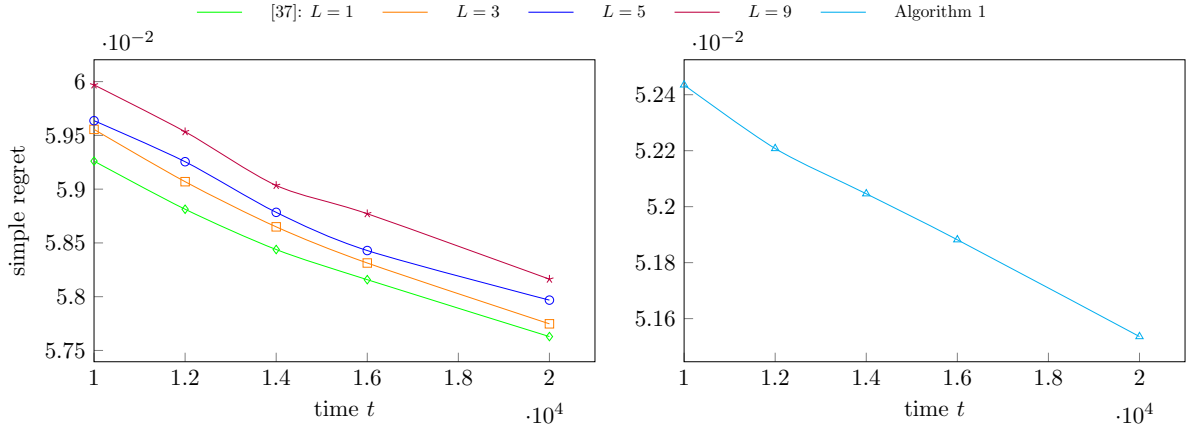

\section{Conclusion}
In conclusion, we have introduced a novel algorithm that operates without prior knowledge of the parameters in the unknown demand function. Beyond its practical applicability, our method also achieves an improved regret upper bound. We have explored extensions of our approach, including handling contextual information and very smooth demand functions.

For future work, establishing the lower bound with respect to $\beta$ remains an open challenge. This would require the careful construction of complex instances to fully understand the limitations of our method.

\newpage

\bibliography{ref}

\newpage
\appendix

\section{Proofs}

\subsection{Proofs in non-contextual model}

\noindent
\textbf{Proof of \cref{prop: invertible gram matrix}.}

\begin{myproof}
  To simplify notation, we denote the prices at rounds in $\mathcal{D}_{t,s}^j$ as $p_1,p_2,\cdots,p_n$ where $n\geq \varpi(\beta)+1$.
We define the matrix
$$
A = [\phi_j(p_1),\phi_j(p_2),\cdots,\phi_j(p_{ \varpi(\beta)+1})].
$$
By the algorithmic construction, the first $ \varpi(\beta)+1$ prices are distinct.
Consequently, $A$ forms a Vandermonde matrix, which is known to be invertible since $\phi_j$ is a polynomial map.

Now, considering all column vectors, we analyze the rank of the set:
$$
\{\phi_j(p_1),\phi_j(p_2),\cdots,\phi_j(p_n)\}.
$$
The rank of this set is $\varpi(\beta)+1$ because the first $\varpi(\beta)+1$ vectors are linearly independent and the degree of the polynomial map $\phi_j$ is also $\varpi(\beta)+1$. 
Thus, we can express the rank of the Gram matrix $ \Lambda_{t,s}^j $ as follows:
\begin{align*}
  rank( \Lambda_{t,s}^j ) =& rank( [\phi_j(p_1),\phi_j(p_2),\cdots,\phi_j(p_n) ]  [\phi_j(p_1)^{\top};\phi_j(p_2)^{\top};\cdots;\phi_j(p_n)^{\top} ] ) \\
  =& rank( [\phi_j(p_1),\phi_j(p_2),\cdots,\phi_j(p_n) ]  ) \\
  =& rank( A  ) \\
  =& \varpi(\beta)+1.
\end{align*}
Since the Gram matrix $\Lambda_{t,s}^j $ is a $(\varpi(\beta)+1)\times(\varpi(\beta)+1)$ matrix, it follows that $\Lambda_{t,s}^j$ is invertible.

\end{myproof}

\noindent
\textbf{Proof of \cref{lem: UCB}.}

\begin{myproof}
  Let $\mathcal{H}_0$ denote the filtration generated by 
  $$
  \left\{ (\tau,p_{\tau}) , \tau \in \bigcup_{ s'\leq s } \Psi_{t}^{s'} \right\} \cup    \left\{ d_{\tau} , \tau \in \bigcup_{ s'< s } \Psi_{t}^{s'} \right\}.
  $$ 
  Based on the algorithmic construction, the selection of round $\tau$ into $\Psi_t^s$ depends solely on $\mathcal{H}_0$. 
  Therefore, condition on $\mathcal{H}_0$, $\{\epsilon_{\tau}, \tau \in \Psi_t^s \}$ are independent random variables.
  Considering the noise $\epsilon_{\tau}$ with index $\tau$ in $\Psi_t^s$, 
  we have $\mathbb{E}[\epsilon_{\tau} | \mathcal{H}_0 ] = 0$ from the realizability condition.
  
  Now we consider the concentration for $\epsilon_{\tau}, \tau\in \mathcal{D}_{t,s}^j$.
  For notation brevity, denote $\mathcal{H}_{\tau}$ as the filtration generated by 
  $
  \{\epsilon_{\tau'}: \tau' \in \mathcal{D}_{t,s}^j, \tau'<\tau\}\cup\mathcal{H}_{0}. 
  $ 
  Then for $\tau=0$, the definition of $\mathcal{H}_{\tau}$ is consistent with $\mathcal{H}_{0}$.
  To apply Azuma's inequality, we need to check $\epsilon_{\tau} $ is a martingale difference sequence adapted to filtration $\mathcal{H}_{\tau}$.
  From the conditional independence of $\epsilon_{\tau}$, we have 
  \begin{align*}
    \mathbb{E}[\epsilon_{\tau}   |\mathcal{H}_{\tau}]   = & \mathbb{E}[\epsilon_{\tau}   |\mathcal{H}_{0}, \{\epsilon_{\tau'}: \tau' \in \mathcal{D}_{t,s}^j, \tau'<\tau\} ] \\
   = &\mathbb{E}[\epsilon_{\tau}   |\mathcal{H}_{0} ] \\
   = & 0.
  \end{align*}

  Let $\tilde{\theta}_{t,s}^j = (\Lambda_{t,s}^j)^{-1}  \sum_{ {\tau}\in\mathcal{D}_{t,s}^j }  [(\theta^j)^{\top} \phi_j(p_{\tau}) + \epsilon_{\tau}  ]   \phi_j(p_{\tau})   $.
  We fix arbitrary $s\in [S], t\in [T]$ and $j\in \mathcal{A}_{t,s}$. 
  From the definition of $\tilde{\theta}_{t,s}^j$, we obtain 
  \begin{align*}
    &  (\tilde{\theta}^j_{t,s}-\theta^j )^{\top} \phi_j(p) \\
  = & \left( (\Lambda_{t,s}^j)^{-1}  \sum_{ {\tau}\in\mathcal{D}_{t,s}^j }  [(\theta^j)^{\top} \phi_j(p_{\tau}) + \epsilon_{\tau}  ]   \phi_j(p_{\tau})  - \theta^j  \right)^{\top} \phi_j(p)   \\
  = &  \phi_j(p)^{\top}   (\Lambda_{t,s}^j)^{-1}  \sum_{ {\tau}\in\mathcal{D}_{t,s}^j }   \epsilon_{\tau}     \phi_j(p_{\tau})  +  \phi_j(p)^{\top} (\Lambda_{t,s}^j)^{-1}  \left( \sum_{ {\tau}\in\mathcal{D}_{t,s}^j }  \phi_j(p_{\tau})\phi_j(p_{\tau})^{\top}  - \Lambda_{t,s}^j   \right)^{\top} \theta^j   \\
  = &  \phi_j(p)^{\top}   (\Lambda_{t,s}^j)^{-1}  \sum_{ {\tau}\in\mathcal{D}_{t,s}^j }   \epsilon_{\tau}     \phi_j(p_{\tau}) .
\end{align*}
  We further apply Azuma's inequality to $\epsilon_{\tau}, \tau\in \mathcal{D}_{t,s}^j$. By doing so, we can derive the following results:
  \begin{align*}
    &\Prob{ |   \phi_j(p)^{\top}   (\Lambda_{t,s}^j)^{-1}  \sum_{ {\tau}\in\mathcal{D}_{t,s}^j }   \epsilon_{\tau}     \phi_j(p_{\tau})   |  \geq  \gamma \sqrt{ \phi_j(p )^{\top}  (\Lambda^j_{t,s})^{-1} \phi_j(p ) }     } \\
  \leq & 2 \exp\left(  - \frac{  2\gamma^2 \phi_j(p )^{\top}  (\Lambda^j_{t,s})^{-1} \phi_j(p )    }{  \sum_{ {\tau}\in\mathcal{D}_{t,s}^j } (  \phi_j(p)^{\top}   (\Lambda_{t,s}^j)^{-1}   \phi_j(p_{\tau})   )^2    }        \right)\\
  = & 2 \exp\left(  - \frac{  2\gamma^2 \phi_j(p )^{\top}  (\Lambda^j_{t,s})^{-1} \phi_j(p )    }{  \phi_j(p)^{\top}   (\Lambda_{t,s}^j)^{-1}  ( \sum_{ {\tau}\in\mathcal{D}_{t,s}^j }    \phi_j(p_{\tau})  \phi_j(p_{\tau})^{\top} )  (\Lambda_{t,s}^j)^{-1} \phi_j(p)  }        \right)\\
  = & 2 \exp\left(  - \frac{ 2 \gamma^2 \phi_j(p )^{\top}  (\Lambda^j_{t,s})^{-1} \phi_j(p )    }{  \phi_j(p)^{\top}   (\Lambda_{t,s}^j)^{-1}     \phi_j(p)  }        \right)\\
  \leq & 2 \exp\left(  -   2 \gamma^2   \right)\\
  = & \frac{\delta}{ NST }.
\end{align*}
  

Therefore, we obtain 
\begin{align*}
  & \Prob{ |(\theta^j -\tilde{\theta}^j_{t,s})^{\top} \phi_j(p) | \geq  \gamma  \sqrt{ \phi_j(p )^{\top}  (\Lambda^j_{t,s})^{-1} \phi_j(p ) }     }  \\  
  \leq  & \Prob{ |   \phi_j(p)^{\top}   (\Lambda_{t,s}^j)^{-1}  \sum_{ {\tau}\in\mathcal{D}_{t,s}^j }   \epsilon_{\tau}     \phi_j(p_{\tau})   |  \geq  \gamma \sqrt{ \phi_j(p )^{\top}  (\Lambda^j_{t,s})^{-1} \phi_j(p ) }     } \\
  \leq & \frac{\delta }{NS T}.
\end{align*}

Moreover, we have
  \begin{align*}
     &  |   (\theta_{t,s}^j -\tilde{\theta}^j_{t,s})^{\top} \phi_j(p)  |        \\
   = & \left| \sum_{ {\tau}\in\mathcal{D}_{t,s}^j }  [(\theta^j)^{\top} \phi_j(p_{\tau}) + \epsilon_{\tau} -d_{\tau} ]   \phi_j(p_{\tau})^{\top} (\Lambda_{t,s}^j)^{-1}  \phi_j(p ) \right|\\
   \leq &  \sum_{ {\tau}\in\mathcal{D}_{t,s}^j }  | (\theta^j)^{\top} \phi_j(p_{\tau}) - f(p_{\tau}) | \cdot |  \phi_j(p_{\tau})^{\top} (\Lambda_{t,s}^j)^{-1}  \phi_j(p ) | \\
  \leq & L(a_{j}-a_{j-1})^{\beta} \sum_{ {\tau}\in\mathcal{D}_{t,s}^j }   | \phi_j(p_{\tau})^{\top} (\Lambda_{t,s}^j)^{-1}  \phi_j(p ) | \\
  = & \frac{L(p_{max}-p_{min})^{\beta}}{N^{\beta}}\sum_{ {\tau}\in\mathcal{D}_{t,s}^j }  |  \phi_j(p_{\tau})^{\top} (\Lambda_{t,s}^j)^{-1}  \phi_j(p )| .
  \end{align*}
  It remains to bound $\sum_{ {\tau}\in\mathcal{D}_{t,s}^j }  p | \phi_j(p_{\tau})^{\top} (\Lambda_{t,s}^j)^{-1}  \phi_j(p ) | $. 
  \begin{align*}
    & \sum_{ {\tau}\in\mathcal{D}_{t,s}^j } p |  \phi_j(p_{\tau})^{\top} (\Lambda_{t,s}^j)^{-1}  \phi_j(p ) |  \\
    \overset{(a)}{\leq} & p \sqrt{ |\mathcal{D}_{t,s}^j |   \sum_{ {\tau}\in\mathcal{D}_{t,s}^j }  |  \phi_j(p_{\tau})^{\top} (\Lambda_{t,s}^j)^{-1}  \phi_j(p ) |^2  } \\
    = & p \sqrt{ |\mathcal{D}_{t,s}^j |    \phi_j(p )(\Lambda_{t,s}^j)^{-1}  \Big(\sum_{ {\tau}\in\mathcal{D}_{t,s}^j }\phi_j(p_{\tau})  \phi_j(p_{\tau})^{\top} \Big) (\Lambda_{t,s}^j)^{-1}  \phi_j(p )   } \\
    = & p \sqrt{ |\mathcal{D}_{t,s}^j |    \phi_j(p )  (\Lambda_{t,s}^j)^{-1}  \phi_j(p )   } \\
    \overset{(b)}{\leq} & 2^{-s} p_{max} \sqrt{ |\mathcal{D}_{t,s}^j |     } \\
    \overset{(c)}{\leq} & p_{max}  \sqrt{ 2 (1+\varpi(\beta))\ln t  },
  \end{align*}
  where the inequality (a) is Cauchy-Schwarz inequality, 
  the inequality (b) is from the condition of the step (a) in \cref{alg: PADP},
  and the inequality (c) is due to \cref{lem: size of dataset j}.

  By the triangular inequality, we find 
  \begin{align*}
    & \Prob{ |p(\theta^j -\theta^j_{t,s})^{\top} \phi_j(p) | \geq  \gamma p \sqrt{ \phi_j(p )^{\top}  (\Lambda^j_{t,s})^{-1} \phi_j(p ) }  + \frac{L p_{max}(p_{max}-p_{min})^{\beta}  \sqrt{2 (1+\varpi(\beta))\ln t} }{ N^{\beta}  }   }  \\  
    \leq  & \mathbb{P} \Bigg( |  p (\theta_{t,s}^j -\tilde{\theta}^j_{t,s})^{\top} \phi_j(p)  |     +  | p  (\theta^j -\tilde{\theta}^j_{t,s})^{\top} \phi_j(p)  |  \\ 
    &  \quad \quad \quad \geq  \gamma p \sqrt{ \phi_j(p )^{\top}  (\Lambda^j_{t,s})^{-1} \phi_j(p ) }  + \frac{Lp_{max}(p_{max}-p_{min})^{\beta}  \sqrt{2 (1+\varpi(\beta))\ln t} }{ N^{\beta}  } \Bigg)  \\
    \leq & \Prob{ |  (\theta^j -\tilde{\theta}^j_{t,s})^{\top} \phi_j(p)  |   \geq  \gamma \sqrt{ \phi_j(p )^{\top}  (\Lambda^j_{t,s})^{-1} \phi_j(p ) }   } \\
    \leq & \frac{\delta }{NS T}.
  \end{align*}

\end{myproof}

\noindent
\textbf{Proof of \cref{lem: difference of best prices for each layer}.}

\begin{myproof}
  We prove this lemma by induction on $s$. 
  For $s=1$, the lemma holds naturally as $\mathcal{A}_{t,1} = [N]$ and thus $rev( p^*   )  =  rev( p_{t,1}^{j^*_{t,1}})  $.
  Assume that the bound holds in the layer $s$.
  It is sufficient to show that 
  $$ 
  rev( p^{j^*_{t,s}}_{t,s} ) - rev( p^{j^*_{t,s+1}}_{t,s+1} )   \leq  \frac{2Lp_{max}(p_{max}-p_{min})^{\beta}  \sqrt{2 (1+\varpi(\beta))\ln t} }{ N^{\beta}  } .
  $$
  If $j_{t,s}^* = j^*_{t,s+1}$, the desired bound holds.
  Hence we assume that $j^*_{t,s} \notin \mathcal{A}_{t,s+1}$.
  Let $\hat{j}_{t,s}:= \argmax_{j\in\mathcal{A}_{t,s}} \sup_{p\in [a_{j-1},a_j]} p U_{t,s}^j (p) $ be the index with the highest UCB in $\mathcal{A}_{t,s}$.
  From the step (a) we know $\hat{j}_{t,s} \in \mathcal{A}_{t,s+1}.$ Then we have 
  \begin{align*}
    & rev( p^{j^*_{t,s}}_{t,s} ) - rev( p^{j^*_{t,s+1}}_{t,s+1} )  \\
   \leq & rev( p^{j^*_{t,s}}_{t,s} ) - rev( p^{\hat{j}_{t,s+1}}_{t,s+1} )   \\
   \leq &  p^{j^*_{t,s}}_{t,s} \phi_{ j^*_{t,s}  } ( p^{j^*_{t,s}}_{t,s} )^{\top}\theta^{j^*_{t,s}}  -p^{\hat{j}_{t,s}}_{t,s} \phi_{ \hat{j}_{t,s}  } ( p^{\hat{j}_{t,s}}_{t,s} )^{\top}\theta^{\hat{j}_{t,s}}  + \frac{2 p_{max} L(p_{max}-p_{min})^{\beta} } { N^{\beta}  } ,
  \end{align*}
  where the last inequality is due to \cref{lem: polynomial regression}.
  From the definition of $\Gamma$, we know that 
  \begin{align*}
   & p^{j^*_{t,s}}_{t,s} \phi_{ j^*_{t,s}  } ( p^{j^*_{t,s}}_{t,s} )^{\top}\theta^{j^*_{t,s}}  -p^{\hat{j}_{t,s}}_{t,s} \phi_{ \hat{j}_{t,s}  } ( p^{\hat{j}_{t,s}}_{t,s} )^{\top}\theta^{\hat{j}_{t,s}}     \\ 
  \leq & p^{j^*_{t,s}}_{t,s} \phi_{ j^*_{t,s}  } ( p^{j^*_{t,s}}_{t,s} )^{\top}\theta_{t,s}^{j^*_{t,s}}+ \gamma p^{j^*_{t,s}}_{t,s} \sqrt{ \phi_{j^*_{t,s}}(p^{j^*_{t,s}}_{t,s})^{\top}  (\Lambda^{j^*_{t,s}}_{t,s})^{-1} \phi_{j^*_{t,s}}(p^{j^*_{t,s}}_{t,s}) }  \\ 
  & \quad \quad    -   p^{\hat{j}_{t,s}}_{t,s} \phi_{ \hat{j}_{t,s}  } ( p^{\hat{j}_{t,s}}_{t,s} )^{\top}\theta_{t,s}^{\hat{j}_{t,s}}  +  \gamma p^{\hat{j}_{t,s}}_{t,s}\sqrt{ \phi_{\hat{j}_{t,s}}(p^{\hat{j}_{t,s}}_{t,s})^{\top}  (\Lambda^{\hat{j}_{t,s}}_{t,s} )^{-1} \phi_{\hat{j}_{t,s}}(p^{\hat{j}_{t,s}}_{t,s}) }   \\ 
  & \quad \quad\quad \quad + \frac{2Lp_{max}(p_{max}-p_{min})^{\beta}  \sqrt{2 (1+\varpi(\beta))\ln t} }{ N^{\beta}  } \\
  = &  p^{j^*_{t,s}}_{t,s} \phi_{ j^*_{t,s}  } ( p^{j^*_{t,s}}_{t,s} )^{\top}\theta_{t,s}^{j^*_{t,s}}+ \gamma p^{j^*_{t,s}}_{t,s} \sqrt{ \phi_{j^*_{t,s}}(p^{j^*_{t,s}}_{t,s})^{\top}  (\Lambda^{j^*_{t,s}}_{t,s})^{-1} \phi_{j^*_{t,s}}(p^{j^*_{t,s}}_{t,s}) }      \\ 
  & \quad \quad -   p^{\hat{j}_{t,s}}_{t,s} \phi_{ \hat{j}_{t,s}  } ( p^{\hat{j}_{t,s}}_{t,s} )^{\top}\theta_{t,s}^{\hat{j}_{t,s}}  - \gamma p^{\hat{j}_{t,s}}_{t,s}\sqrt{ \phi_{\hat{j}_{t,s}}(p^{\hat{j}_{t,s}}_{t,s})^{\top}  (\Lambda^{\hat{j}_{t,s}}_{t,s} )^{-1} \phi_{\hat{j}_{t,s}}(p^{\hat{j}_{t,s}}_{t,s}) }   \\ 
  & \quad \quad\quad \quad + 2\gamma p^{\hat{j}_{t,s}}_{t,s}\sqrt{ \phi_{\hat{j}_{t,s}}(p^{\hat{j}_{t,s}}_{t,s})^{\top}  (\Lambda^{\hat{j}_{t,s}}_{t,s} )^{-1} \phi_{\hat{j}_{t,s}}(p^{\hat{j}_{t,s}}_{t,s}) }   + \frac{2Lp_{max}(p_{max}-p_{min})^{\beta}  \sqrt{2 (1+\varpi(\beta))\ln t} }{ N^{\beta}  }\\
 =  & p^{j^*_{t,s}}_{t,s}  U_{t,s}^{j^*_{t,s}}( p^{j^*_{t,s}}_{t,s} )  - p^{\hat{j}_{t,s}}_{t,s}   U_{t,s}^{\hat{j}_{t,s}}( p^{\hat{j}_{t,s}}_{t,s}  )   \\ 
 & \quad \quad + 2\gamma p^{\hat{j}_{t,s}}_{t,s}\sqrt{ \phi_{\hat{j}_{t,s}}(p^{\hat{j}_{t,s}}_{t,s})^{\top}  (\Lambda^{\hat{j}_{t,s}}_{t,s} )^{-1} \phi_{\hat{j}_{t,s}}(p^{\hat{j}_{t,s}}_{t,s}) }   + \frac{2Lp_{max}(p_{max}-p_{min})^{\beta}  \sqrt{2 (1+\varpi(\beta))\ln t} }{ N^{\beta}  }.
\end{align*}

From the algorithmic construction (step (b)), we know 
$$
\gamma p^{\hat{j}_{t,s}}_{t,s}\sqrt{ \phi_{\hat{j}_{t,s}}(p^{\hat{j}_{t,s}}_{t,s})^{\top}  (\Lambda^{\hat{j}_{t,s}}_{t,s} )^{-1} \phi_{\hat{j}_{t,s}}(p^{\hat{j}_{t,s}}_{t,s}) }  \leq p_{max}2^{-s}.
$$
Since $ {j}^*_{t,s} \notin \mathcal{A}_{t,s} $ by our assumption, the step (a) yields 
\begin{align*}
  & p^{\hat{j}_{t,s}}_{t,s}   U_{t,s}^{\hat{j}_{t,s}}( p^{\hat{j}_{t,s}}_{t,s}  )   - p^{j^*_{t,s}}_{t,s}  U_{t,s}^{j^*_{t,s}}( p^{j^*_{t,s}}_{t,s} )\\
 \geq &  p^{\hat{j}_{t,s}}_{t,s}   U_{t,s}^{\hat{j}_{t,s}}( p^{\hat{j}_{t,s}}_{t,s}  )   -   \sup_{p\in [a_{j^*_{t,s} -1}, a_{j^*_{t,s} }]} pU_{t,s}^{j^*_{t,s}}(p) \\
  >  & p_{max}2^{1-s} \\ 
\geq & 2 \gamma p^{\hat{j}_{t,s}}_{t,s}\sqrt{ \phi_j(p^{\hat{j}_{t,s}}_{t,s})^{\top}  \Lambda^j_{t,s} \phi_j(p^{\hat{j}_{t,s}}_{t,s}) }.
\end{align*}

Combing all inequalities above, we obtain 
\begin{align*}
  rev( p^{j^*_{t,s}}_{t,s} ) - rev( p^{j^*_{t,s+1}}_{t,s+1} )   \leq  \frac{2Lp_{max}(p_{max}-p_{min})^{\beta}  \sqrt{2 (1+\varpi(\beta))\ln t} }{ N^{\beta}  } .
\end{align*}

\end{myproof}

\noindent
\textbf{Proof of \cref{lem: regret of prices at layer s}.}

\begin{myproof}
  For any layer $s < s_t$, the step (a) shows that 
  \begin{align*}
    p_{max} 2^{1-s} \geq  & \max \limits_{  j' \in \mathcal{A}_{t,s} } \sup_{p\in [a_{j-1},a_j] } p U^{j'}_{t,s}(p)   -  \sup_{ p \in [a_{j_t-1},a_{j_t}] } p  U^{j_t}_{t,s}(p)  \\
    \geq   &  p^{j^*_{t,s}}_{t,s} U^{j^*_{t,s}}_{t,s}(p^{j^*_{t,s}}_{t,s})   -  p_t \phi_{j_t}(p_t)^{\top} \theta^{ j_t  }  +  \frac{Lp_{max}(p_{max}-p_{min})^{\beta}  \sqrt{2 (1+\varpi(\beta))\ln t} }{ N^{\beta}  },
  \end{align*}
  where the last inequality is due to the definition of $\Gamma$.
  
  The step (b) implies that 
  $$
  \sup_{  p \in [a_{j_t-1},a_{j_t}]  }  \gamma p  \sqrt{ \phi_{j_t}(p )^{\top}  ( \Lambda^{j_t}_{t,s})^{-1}\phi_{j_t}(p ) } \leq   p_{max}2^{1-s}
  $$
  as $j_t\in\mathcal{A}_{t,s}$ and $s<s_t$. 
  Combing two inequalities we obtain 
  \begin{align*}
  &  p^{j^*_{t,s}}_{t,s} U_{ j^*_{t,s}  } ( p^{j^*_{t,s}}_{t,s} )   -\sup_{p \in [a_{j_{t}-1},a_{j_t}]}  p \bigg(U_{t,s}^{j_t}(p) -   2\gamma   \sqrt{ \phi_{j_t}(p)^{\top}  ( \Lambda^{j_t}_{t,s})^{-1}\phi_{j_t}(p) } \bigg)   \\
  \leq & 2 \sup_{p \in [a_{j_{t}-1},a_{j_t}]} \gamma p  \sqrt{ \phi_{j_t}(p)^{\top}  ( \Lambda^{j_t}_{t,s})^{-1}\phi_{j_t}(p) }  +  p_{max} 2^{1-s} \\
  \leq & 3 p_{max} 2^{1-s}.
  \end{align*}

  Therefore, from the definition of $\Gamma$ we have 
  \begin{align*}
     &  3 p_{max} 2^{1-s}    \\
  \geq & p^{j^*_{t,s}}_{t,s} U_{ j^*_{t,s}  } ( p^{j^*_{t,s}}_{t,s} )   -\sup_{p \in [a_{j_{t}-1},a_{j_t}]}  p \bigg(U_{t,s}^{j_t}(p) -   2\gamma   \sqrt{ \phi_{j_t}(p)^{\top}  ( \Lambda^{j_t}_{t,s})^{-1}\phi_{j_t}(p) } \bigg)   \\
  \geq & p^{j^*_{t,s}}_{t,s} \phi_{ j^*_{t,s}  } ( p^{j^*_{t,s}}_{t,s} )^{\top}\theta^{j^*_{t,s}}   - \sup_{p\in[a_{j_t-1},a_{j_t}]} p \phi_{j_t}(p)^{\top} \theta^{ j_t  }     - \frac{2Lp_{max}(p_{max}-p_{min})^{\beta}  \sqrt{2 (1+\varpi(\beta))\ln t} }{ N^{\beta}  }   .
  \end{align*}
 Rearranging all terms yields the desired inequality.

\end{myproof}

\noindent
\textbf{Proof of \cref{lem: discrete regret at the round t}.}

\begin{myproof}
  It follows from \cref{lem: polynomial regression} and \cref{lem: regret of prices at layer s} that 
  \begin{align*}
   & rev( p^*   ) - rev( p_t ) \\
   \leq &   rev( p^*   ) - rev( p_{t,s_t-1}^{j^*_{t,s_t-1}})  +  rev( p_{t,s_t-1}^{j^*_{t,s_t-1}}) -  rev( p_t )  \\
   \leq &  \frac{2 Lp_{max}(p_{max}-p_{min})^{\beta}  \sqrt{2 (1+\varpi(\beta))\ln t} }{ N^{\beta}  }    (s_t-2)+  rev( p_{t,s_t-1}^{j^*_{t,s_t-1}}) -  rev( p_t )  \\
   \leq & \frac{Lp_{max}(p_{max}-p_{min})^{\beta}  \sqrt{2 (1+\varpi(\beta))\ln t} }{ N^{\beta}  }    (2s_t-2)+  p^{j^*_{t,s_t-1}}_{t,s_t-1} \phi_{ j^*_{t,s_t-1}  } ( p^{j^*_{t,s_t-1}}_{t,s_t-1} )^{\top}\theta^{j^*_{t,s_t-1}}  -  p_t \phi_{ j_t  } ( p_t )^{\top}\theta^{j_t}  \\
   \leq & \frac{Lp_{max}(p_{max}-p_{min})^{\beta}  \sqrt{2 (1+\varpi(\beta))\ln t} }{ N^{\beta}  }    (2s_t-2) \\ 
   & \quad \quad \quad + 6p_{max} \cdot 2^{1-s_t} +   \frac{Lp_{max}(p_{max}-p_{min})^{\beta}  \sqrt{2 (1+\varpi(\beta))\ln t} }{ N^{\beta}  } \\
   \leq& 12 p_{max} \cdot   2^{-s_t}+ \frac{Lp_{max}(p_{max}-p_{min})^{\beta}  \sqrt{2 (1+\varpi(\beta))\ln t} }{ N^{\beta}  }    (2s_t-1).
  \end{align*}

\end{myproof}

\noindent
\textbf{Proof of \cref{thm: regret upper bound}.}

\begin{myproof}
  Let $\Psi_0$ be the set of rounds for which an alternative is chosen when $   p_t \sqrt{ \phi_j(p_t )^{\top}  (\Lambda^j_{t,1})^{-1} \phi_j(p_t ) }     \leq 1/\sqrt{T}$.
  Since $2^{-S} \leq 1/\sqrt{T}$, we have $ \Psi_0 \cup \cup_{s\in [S]}\Psi^s_{T+1} = [T]$. 
  
  Recall that $p^*$ maximizes $pf(p)$.
   \begin{align*}
      & \mathbb{E}[  Reg(T) ] \\
      = &   \sum_{t=1   }^{T}  \mathbb{E}\left[ p^*f(p^*) - p_tf(p_t)     \right] \\ 
      = & \sum_{s=2}^S \sum_{t\in \Psi_{T +1}^s } \mathbb{E}\Big[   p^*f(p^*) - p_tf(p_t)    \Big] + \sum_{t\in \Psi_{T +1}^1 } \mathbb{E}\Big[   p^*f(p^*) - p_tf(p_t)    \Big] + \sum_{t\in \Psi_0 } \mathbb{E}\Big[   p^*f(p^*) - p_tf(p_t)    \Big] \\ 
    \leq&  \sum_{s=2}^S \left( 12 p_{max} \cdot 2^{-s}+ \frac{Lp_{max}(p_{max}-p_{min})^{\beta}  \sqrt{2 (\varpi(\beta) +1)\ln T} }{ N^{\beta}  }    (2s-1) \right)  |\Psi_{T+1}^s| \\
    &  \quad \quad\quad\quad +p_{max}|\Psi_{T+1}^1|    +  \sum_{t\in \Psi_0 } \mathbb{E}\Big[   p^*f(p^*) - p_tf(p_t)    \Big]     ,
  \end{align*}
  where the last inequality is due to \cref{lem: discrete regret at the round t}.

 From \cref{lem: size of dataset j}, we have 
 $$
 |\Psi_{T +1}^s| \leq 2^s \gamma \sqrt{2N (\varpi(\beta)+1 )  | \Psi_{T+1}^s| \ln T },
 $$
 which implies 
 $$
 |\Psi_{T +1}^s| \leq 2^{2s+1} \gamma^2 N (\varpi(\beta) +1) \ln T.
 $$
 As a corollary, we have 
 $$
 p_{max}|\Psi_{T+1}^1| \leq 8 p_{max} N (\varpi(\beta) +1) \ln^2 (T/\delta).
 $$

 Therefore, we have 
 \begin{align*}
  \sum_{s=2}^S 12 p_{max} \cdot 2^{-s}|\Psi_{T +1}^s| \leq & \sum_{s=2}^S  12p_{max}  \sqrt{2N (\varpi(\beta) +1) | \Psi_{T+1}^s| \ln T } \\
 & \leq  12 \gamma p_{max}\sqrt{2N (\varpi(\beta) +1)(S-1)\sum_{s=2}^S  | \Psi_{T+1}^s| \ln T } \\
 & \leq  12 \gamma p_{max}\sqrt{N (\varpi(\beta) +1) {T}  \ln T  \log_2 T    }
  \end{align*}
 and
 \begin{align*}
   & \sum_{s=2}^S   \frac{Lp_{max}(p_{max}-p_{min})^{\beta}  \sqrt{2 (\varpi(\beta) +1) \ln T} }{ N^{\beta}  }    (2s-1)   |\Psi_{T+1}^s| \\ 
  \leq & \frac{2 T Lp_{max}(p_{max}-p_{min})^{\beta}  \sqrt{2 (\varpi(\beta) +1) \ln T} }{ N^{\beta}  }   \log_2 T. 
 \end{align*}
 
 From \cref{lem: regret at the round t in Psi0}, we know that 
$$
\sum_{t\in \Psi_0 } \mathbb{E}\Big[   p^*f(p^*) - p_tf(p_t)    \Big] \leq 2p_{max} \sqrt{T} +  \frac{2Lp_{max}(p_{max}-p_{min})^{\beta}  }{ N^{\beta}  } .
$$

For the regret caused in the initialization step,
we directly bound the caused regret as 
$$
N(1+\varpi(\beta))p_{max}.
$$

  Since $N = \lceil  (p_{max}-p_{min})^{\frac{2\beta}{2\beta+1}}   T^{\frac{1}{2\beta+1}}     \rceil$, we have 
  \begin{align*}
    & \mathbb{E}[Reg(T)] \leq (12+  L  )p_{max} (p_{max}-p_{min})^{\frac{\beta}{2\beta+1}}   (\varpi(\beta) +1)^{\frac{1}{2 }}  \log^{\frac{3}{2}}_2 (T/\delta ) T^{ \frac{\beta+1}{2\beta+1} }  \\ 
    & \quad \quad \quad +  2p_{max} \sqrt{T} + 17 p_{max}  (p_{max}-p_{min})^{\frac{2\beta}{2\beta+1}} (\varpi(\beta) +1)^{\frac{3}{2 }} ( T \log_2 T )^{\frac{1}{2\beta+1}}     \ln T   .
\end{align*}
\end{myproof}

\noindent
\textbf{Proof of \cref{lem: stopping time}.}
\begin{myproof}
  Consider the interval $[a_{j-1},a_j]$.
  We define the following stopping time 
  $$
  \mathcal{T}_j = \min\{ n\leq T| \lambda_{min}(\Lambda_n^j)\geq 1  \},
  $$
  where $n$ denotes the sample size of $\mathcal{D}^j$.
  Consequently, we have
  $$
  \mathcal{T} \leq \sum_{j=1}^N \mathcal{T}_j.
  $$

  Now we apply \cref{lem: eigenvalue} to establish an upper bound for $\mathcal{T}_j$.
  We have 
  $$
  \mathbb{E}[\mathcal{T}_j ] \leq  \left( \frac{C_1\sqrt{\alpha + \varpi(\beta)}+C_2\sqrt{\log(1/\delta)}}{\lambda_{min}(\Sigma)}    \right)^2 + \frac{2}{\lambda_{min}(\Sigma)} + \delta T
  $$
  for $\delta >0$.
  By setting $\delta=1/T$ we can prove $\mathbb{E}[\mathcal{T}_j ] = \mathcal{O}(\log T) $.
  Finally, we conclude the proof by summing over the $N$ stopping times, yielding the desired result.
  
\end{myproof}

\noindent
\textbf{Proof of \cref{lem: estimation upper bound}.}

\begin{myproof}
  Let $\mathbf{P}_n$ be a $n\times \ell$ matrix with its $j$-th row $\varphi(p_j,x_j)^\top $ for every $m$ and $\mathbf{d}_n =(d_1,d_2,\cdots,d_n)^{\top}$.
  By the least square regression, we obtain 
  $$
  \hat{\theta} = (\mathbf{P}_n^\top \mathbf{P}_n)^{-1} \mathbf{P}_n^{\top} \mathbf{d}_n.
  $$
  Define 
  $$
  \theta_0 =  \mathbb{E}[ \varphi(p_1,x_1) \varphi(p_1,x_1)^\top ]^{-1} \mathbb{E}[  \varphi(p_1,x_1) d_1 ]  
  $$
  According to least square estimation theory \citep{SmoothnessApaptiveDP}, we have
  $$
  \norm{\hat{\theta} -\theta_0 } < n^{-\frac{1}{2}(1-v)}\ln^3 n
  $$
  with probability at least $1-O(\ell e^{-C_2\ln^2 n})$ for some constants $C_2$ depending on $u_2$, $\alpha$ and $\beta$,
  and for $n$ larger than some constants $C_1$ depending on $u_1'$, $u_2$, $\ell$.

  Note that 
  $$
  \Gamma_{\ell}^{[a,b]\times \mathcal{X}}\{\mathbb{E}[d_1|p_1,x_1]\} =  \Gamma_{\ell}^{[a,b]}f(p) + \mu^\top x = \varphi (p,x)^\top \theta_0   .
  $$

  By the H{\"o}lder assumption and \cref{lem: polynomial regression}, there exists an $\ell+1$ dimensional vector $\theta_1$ such that 
  $$
  |   \mathbb{E}[ \mathbb{E}[ d_1|p_1,x_1  ]  -  \varphi(p,x)^\top \theta_1] | = \mathcal{O}((b-a)^\beta).
  $$
  for $\forall p\in [a,b]$.
  Hence, we have 
  \begin{align*}
     \norm{\theta_0-\theta_1} &= \Big\| \mathbb{E}[ \varphi(p_1,x_1) \varphi(p_1,x_1)^\top ]^{-1} \mathbb{E}[  \varphi(p_1,x_1) d_1 ]   \\
     &\quad \quad - \mathbb{E}[ \varphi(p_1,x_1) \varphi(p_1,x_1)^\top ]^{-1} \mathbb{E}[ \varphi(p_1,x_1) \varphi(p_1,x_1)^\top ] \theta_1 \Big\| \\
     & = \norm{  \mathbb{E}[ \varphi(p_1,x_1) \varphi(p_1,x_1)^\top ]^{-1} \mathbb{E}[ \varphi(p_1,x_1) (   d_1 - \varphi(p_1,x_1)^\top \theta_1   )    ]    } \\
     & = \mathcal{O}((b-a)^{\beta}).
  \end{align*}
  Therefore, we obtain that 
  $$
 \norm{\hat{\theta}-\theta_1} \leq n^{-\frac{1}{2}(1-v)}+(b-a)^{\beta} \ln n,
  $$
  for some larger $n>C_1$ depending on $u_1'$, $u_2$, $\ell$.
  
  Note that $\varphi(p,x) = \mathcal{O}(1)$ and $\norm{\varphi(p,x)^\top(\hat{\theta}-\theta_1) } = \mathcal{O}(\norm{\hat{\theta} -\theta_1})$, $\forall p\in[a,b]$, $n>C_1$.
  We have 
  $$
  | \mathbb{E}[d_1|p_1,x_1] - \varphi(p_1,x_1)^\top \hat{\theta}  | \leq n^{-\frac{1}{2}(1-v)}+(b-a)^{\beta} \ln n.
  $$
  We conclude the proof by noticing that $\mathbb{E}[d|p,x] = f(p)+\mu^\top x$.
\end{myproof}

\noindent
\textbf{Proof of \cref{thm: beta confidence bound}.}
\begin{myproof}
  We first define an event $A=\{ \exists i\in \{1,2\}, m\in\{1,2,\cdots,K_i\}, s.t., |O_{i,m}|<\frac{T_i}{2K_i}  \}$.
  From the concentration inequality, we have 
  $$
  \mathbb{P}(A) \leq T \left( exp(-\frac{T_1}{50K_1}) + exp(-\frac{T_2}{50K_2})  \right).
  $$
  Bt conditioning on $A^c$, we can guarantee the number of samples in each interval. 
  
  We then consider the inequalities when the event $A^c$ holds.
  Define the interval $\mathbf{I}_{i,m} = [p_{min}+\frac{(m-1)(p_{max}-p_{min})}{K_i},p_{min}+\frac{m(p_{max}-p_{min})}{K_i}  ]$.
  Invoking \cref{lem: estimation upper bound}, with a probability of at least $1-\mathcal{O}(\ell e^{-C\ln^2 T})$, we have $\forall p\in \mathbf{I}_{i,m}$,
  $$
  |f(p)+\mu^\top x - \hat{f}_i(p) - \hat{\mu}_i^\top x| < K_i^{-\beta} \ln T + (\frac{T_i}{2K_i})^{-\frac{1}{2} (1-v_i)}\ln^3 T,
  $$
  for some sufficiently small constants $v_1,v_2$.
  Subsequently, we deduce the following upper bound 
  \begin{align*}
    & |\hat{f}_1(p)+\hat{\mu}_1^\top x - \hat{f}_2(p)-\hat{\mu}_2(x)| \\
    \leq & | f(p) +\mu^x - \hat{f}_1(p)-\hat{\mu}_1^\top x | + | f(p) +\mu^x - \hat{f}_2(p)-\hat{\mu}_2^\top x | \\
     \leq & (K_1+K_2)^{-\beta} \ln T + \ln^3 T \left[  (\frac{T}{2K_1})^{ -\frac{1}{2}(1-v_1) } +(\frac{T}{2K_2})^{ -\frac{1}{2}(1-v_2) }  \right] \\
     \leq & 2^c T^{-\beta k_2}\ln T.
  \end{align*}
  for a small constant $c>0$.
  
  From the proof of \cref{lem: estimation upper bound}, we have 
  $$
  |\Gamma_{\ell}^{ \mathbf{I}_{i,m} } f(p) + \mu^\top x - \hat{f}_2(p) -\hat{\mu}_{2}^\top x| = | \varphi(p,x)^\top (\hat{\theta} - \theta_0) | \leq  (\frac{T_2}{2K_2})^{-\frac{1}{2}(1-v_2)}\ln^3 T.
  $$
  
  Given the self-similarity condition of $f$f, we can establish 
  $$
  \norm{f- \Gamma_{\ell}^{ \mathbf{I}_{i,m} } f} \geq M_2 K_2^{-\beta}.
  $$
  
  Combining all inequalities above, a lower bound can be derived as
  \begin{align*}
    & |\hat{f}_1(p)+\hat{\mu}_1^\top x - \hat{f}_2(p)-\hat{\mu}_2^\top x| \\
    \geq & \norm{f- \Gamma_{\ell}^{ \mathbf{I}_{i,m} } f} - |f(p)+\mu^\top x - \hat{f}_1(p) - \hat{\mu}_1^\top x| - |\Gamma_{\ell}^{ \mathbf{I}_{i,m} } f(p) + \mu^\top x - \hat{f}_2(p) -\hat{\mu}_{2}^\top x| \\
    \geq & M_2 K_2^{-\beta} - K_1^{-\beta} \ln T - \ln^3 T \left[  (\frac{T}{2K_1})^{ -\frac{1}{2}(1-v_1) } +(\frac{T}{2K_2})^{ -\frac{1}{2}(1-v_2) }  \right] \\
    \geq & \frac{M_2}{2} T^{-\beta k_2}.
  \end{align*}
  
  From the algorithmic construction, we have 
  \begin{align*}
     \hat{\beta}  & = -\frac{\ln( \max_{p,x}{|\hat{f}_2+\hat{\mu}_2^\top x-\hat{f}_1-\hat{\mu}_1^\top x|}  )}{\ln T} - \frac{\ln \ln T}{\ln T} \\
     & \in \left[ \beta - \frac{c\ln2+\ln\ln T}{ k_2\ln T} - \frac{\ln \ln T}{\ln T} , \beta - \frac{\ln(M_2/2)}{k_2 \ln T} - \frac{\ln \ln T}{\ln T}   \right] \\
     & \subset \left[ \beta - \frac{4(\beta_{max}+1)\ln\ln T}{ \ln T}  , \beta    \right]
  \end{align*}
  with probability at least $1-\mathcal{O}((\alpha+\varpi(\beta_{max}) +1 ) e^{-C\ln^2 T})$.

  \end{myproof}

\begin{mylem}
   \label{lem: bound of r}
   Assuming $|\mathcal{D}_{t,s}^j|\geq 2$, then for arbitrary $t$ and $s$, we have
   $$
   \sum_{\tau\in \mathcal{D}_{t,s}^j} p_{\tau}  \gamma  \sqrt{ \phi_j(p_{\tau})^{\top}  (\Lambda^j_{\tau,s})^{-1} \phi_j(p_{\tau}) }     \leq \gamma p_{max} \sqrt{2 (1+\varpi(\beta)) | \mathcal{D}_{t,s}^j| \ln | \mathcal{D}_{t,s}^j|  }.
   $$
\end{mylem}

\begin{myproof}
   We have
   $$
   \sum_{\tau\in \mathcal{D}_{t,s}^j} p_{\tau}  \gamma  \sqrt{ \phi_j(p_{\tau})^{\top}  (\Lambda^j_{\tau,s})^{-1} \phi_j(p_{\tau}) } \leq  \gamma   p_{max}  \sum_{\tau\in \mathcal{D}_{t,s}^j}    \sqrt{ \phi_j(p_{\tau})^{\top}  (\Lambda^j_{\tau,s})^{-1} \phi_j(p_{\tau}) }
   $$
   From the lemma 13 in \citep{supLinUCB} or the lemma 3 in \citep{linContextual}, we know that 
   $$
   \sum_{\tau\in \mathcal{D}_{t,s}^j}    \sqrt{ \phi_j(p_{\tau})^{\top}  (\Lambda^j_{\tau,s})^{-1} \phi_j(p_{\tau}) } \leq \sqrt{2 (1+\varpi(\beta)) | \mathcal{D}_{t,s}^j| \ln | \mathcal{D}_{t,s}^j|  }
   $$

\end{myproof}

\begin{mylem}
   \label{lem: size of dataset j}
   For all $s$, we have 
   $$
   |\mathcal{D}_{t,s}^j | \leq 2^s \gamma \sqrt{2 (1+\varpi(\beta)) | \mathcal{D}_{t,s}^j| \ln | \mathcal{D}_{t,s}^j|  }.
   $$
\end{mylem}

\begin{myproof}
   From the step (b), we know that 
   $$
   \sum_{\tau\in \mathcal{D}_{t,s}^j} p_{\tau}  \gamma  \sqrt{ \phi_j(p_{\tau})^{\top}  (\Lambda^j_{\tau,s})^{-1} \phi_j(p_{\tau}) }    \geq p_{max} 2^{-s} |\mathcal{D}_{t,s}^j |.
   $$
   By \cref{lem: bound of r}, we obtain 
   $$
   \sum_{\tau\in \mathcal{D}_{t,s}^j} p_{\tau}  \gamma  \sqrt{ \phi_j(p_{\tau})^{\top}  (\Lambda^j_{\tau,s})^{-1} \phi_j(p_{\tau}) }    \leq  \gamma p_{max} \sqrt{2 (1+\varpi(\beta)) | \mathcal{D}_{t,s}^j| \ln | \mathcal{D}_{t,s}^j|  }.
   $$
   Therefore, combing above inequalities, we have 
   $$
   |\mathcal{D}_{t,s}^j | \leq 2^s \gamma \sqrt{2 (1+\varpi(\beta)) | \mathcal{D}_{t,s}^j| \ln | \mathcal{D}_{t,s}^j|  }.
   $$

\end{myproof}

\begin{mylem}
   \label{lem: regret at the round t in Psi0}
   Given the event $\Gamma$, for every round $t\in \Psi_0$, we have 
   $$
   rev( p^*   ) - rev( p_t ) \leq \frac{2p_{max}}{\sqrt{T}}  +  \frac{2Lp_{max}(p_{max}-p_{min})^{\beta}    }{ N^{\beta}  } .
   $$
 \end{mylem}
 \begin{myproof}
   Let $j^*$ be the index such that $p^*\in [a_{{j^*}-1},a_{{j^*}}]$.
   \begin{align*}
    &  rev( p^*   ) - rev( p_t ) \\
    \leq & p^* U^{j^*}_{t,1}(p^*) - p_t U^{j_t}_{t,1}(p_t)  +  \frac{2p_{max}}{\sqrt{T}}  +  \frac{2Lp_{max}(p_{max}-p_{min})^{\beta}    }{ N^{\beta}  }  \\ 
    \leq & \frac{2p_{max}}{\sqrt{T}}  +  \frac{2Lp_{max}(p_{max}-p_{min})^{\beta}    }{ N^{\beta}  }  \\ 
   \end{align*}
 \end{myproof}

\subsection{Proofs in linear contextual effect model}

\begin{mylem}
   \label{lem: UCB in linear contextual effect setting}
   Consider the round $t$ and any layer $s\in [s_t]$ in the round $t$.
   Then for each $j\in[N]$, we have 
   $$
   |p(\theta^j -\theta^j_{t,s})^{\top} \varphi_j(p,x) | \leq  \gamma p\sqrt{ \varphi_j(p,x)^{\top}  (\Lambda^j_{t,s})^{-1} \varphi_j(p,x) }  + \frac{Lp_{max}(p_{max}-p_{min})^{\beta}  \sqrt{2 (\alpha+\varpi(\beta)+1)\ln t} }{ N^{\beta}  }   ,
   $$
   with probability at least $ 1- \frac{\delta }{NS T}$.
\end{mylem}

\begin{myproof}
  In this proof, we modify the filtration $\mathcal{H}_0$ as the following one generated by
  $$
  \left\{ (\tau,x_{\tau},p_{\tau}) , \tau \in \bigcup_{ s'\leq s } \Psi_{t}^{s'} \right\} \cup    \left\{ d_{\tau} , \tau \in \bigcup_{ s'< s } \Psi_{t}^{s'} \right\}.
  $$ 
  The remaining argument is the same as that in \cref{lem: UCB}.

   Let $\tilde{\theta}_{t,s}^j = (\Lambda_{t,s}^j)^{-1}  \sum_{ {\tau}\in\mathcal{D}_{t,s}^j }  [(\theta^j)^{\top} \varphi_j(p_{\tau},x_{\tau}) + \epsilon_{\tau}  ]   \varphi_j(p_{\tau},x_{\tau})   $.
   We fix arbitrary $s\in [S], t\in [T]$ and $j\in \mathcal{A}_{t,s}$. 
   From the definition of $\tilde{\theta}_{t,s}^j$, we obtain 
   \begin{align*}
     &  (\tilde{\theta}^j_{t,s}-\theta^j )^{\top} \varphi_j(p,x) \\
   = & \left( (\Lambda_{t,s}^j)^{-1}  \sum_{ {\tau}\in\mathcal{D}_{t,s}^j }  [(\theta^j)^{\top} \varphi_j(p_{\tau},x_{\tau}) + \epsilon_{\tau}  ]   \varphi_j(p_{\tau},x_{\tau})  - \theta^j  \right)^{\top} \varphi_j(p,x)   \\
   = &  \varphi_j(p,x)^{\top}   (\Lambda_{t,s}^j)^{-1}  \sum_{ {\tau}\in\mathcal{D}_{t,s}^j }   \epsilon_{\tau}     \varphi_j(p_{\tau},x_{\tau})  +  \varphi_j(p,x)^{\top} (\Lambda_{t,s}^j)^{-1}  \left( \sum_{ {\tau}\in\mathcal{D}_{t,s}^j }  \varphi_j(p_{\tau},x_{\tau})\varphi_j(p_{\tau},x_{\tau})^{\top}  - \Lambda_{t,s}^j   \right)^{\top} \theta^j   \\
   = &  \varphi_j(p,x)^{\top}   (\Lambda_{t,s}^j)^{-1}  \sum_{ {\tau}\in\mathcal{D}_{t,s}^j }   \epsilon_{\tau}     \varphi_j(p_{\tau},x_{\tau}) .
 \end{align*}
   We further apply Azuma's inequality to $\epsilon_t$. By doing so, we can derive the following results:
   \begin{align*}
     &\Prob{ |   \varphi_j(p,x)^{\top}   (\Lambda_{t,s}^j)^{-1}  \sum_{ {\tau}\in\mathcal{D}_{t,s}^j }   \epsilon_{\tau}     \varphi_j(p_{\tau},x_{\tau})   |  \geq  \gamma \sqrt{ \varphi_j(p,x)^{\top}  (\Lambda^j_{t,s})^{-1} \varphi_j(p,x) }     } \\
   \leq & 2 \exp\left(  - \frac{  2\gamma^2 \varphi_j(p,x)^{\top}  (\Lambda^j_{t,s})^{-1} \varphi_j(p,x)    }{  \sum_{ {\tau}\in\mathcal{D}_{t,s}^j } (  \varphi_j(p,x)^{\top}   (\Lambda_{t,s}^j)^{-1}   \varphi_j(p_{\tau},x_{\tau})   )^2    }        \right)\\
   = & 2 \exp\left(  - \frac{  2\gamma^2 \varphi_j(p,x)^{\top}  (\Lambda^j_{t,s})^{-1} \varphi_j(p,x)    }{  \varphi_j(p,x)^{\top}   (\Lambda_{t,s}^j)^{-1}  ( \sum_{ {\tau}\in\mathcal{D}_{t,s}^j }    \varphi_j(p_{\tau},x_{\tau})  \varphi_j(p_{\tau},x_{\tau})^{\top} )  (\Lambda_{t,s}^j)^{-1} \varphi_j(p,x)  }        \right)\\
   = & 2 \exp\left(  - \frac{ 2 \gamma^2 \varphi_j(p,x)^{\top}  (\Lambda^j_{t,s})^{-1} \varphi_j(p,x)    }{  \varphi_j(p,x)^{\top}   (\Lambda_{t,s}^j)^{-1}     \varphi_j(p,x)  }        \right)\\
   \leq & 2 \exp\left(  -   2 \gamma^2   \right)\\
   = & \frac{\delta}{ NST }.
 \end{align*}

 Therefore, we obtain 
 \begin{align*}
   & \Prob{ |(\theta^j -\tilde{\theta}^j_{t,s})^{\top} \varphi_j(p,x) | \geq  \gamma  \sqrt{ \varphi_j(p,x)^{\top}  (\Lambda^j_{t,s})^{-1} \varphi_j(p,x) }     }  \\  
   \leq  & \Prob{ |   \varphi_j(p,x)^{\top}   (\Lambda_{t,s}^j)^{-1}  \sum_{ {\tau}\in\mathcal{D}_{t,s}^j }   \epsilon_{\tau}     \varphi_j(p_{\tau},x_{\tau})   |  \geq  \gamma \sqrt{ \varphi_j(p,x)^{\top}  (\Lambda^j_{t,s})^{-1} \varphi_j(p,x) }     } \\
   \leq & \frac{\delta }{NS T}.
 \end{align*}

 Moreover, we have
   \begin{align*}
      &  |   (\theta_{t,s}^j -\tilde{\theta}^j_{t,s})^{\top} \varphi_j(p,x)  |        \\
    = & \left| \sum_{ {\tau}\in\mathcal{D}_{t,s}^j }  [(\theta^j)^{\top} \varphi_j(p_{\tau},x_{\tau}) + \epsilon_{\tau} -d_{\tau} ]   \varphi_j(p_{\tau},x_{\tau})^{\top} (\Lambda_{t,s}^j)^{-1}  \varphi_j(p,x) \right|\\
    \leq &  \sum_{ {\tau}\in\mathcal{D}_{t,s}^j }  | (\theta^j)^{\top} \varphi_j(p_{\tau},x_{\tau}) - f(p_{\tau}) | \cdot |  \varphi_j(p_{\tau},x_{\tau})^{\top} (\Lambda_{t,s}^j)^{-1}  \varphi_j(p,x) | \\
   \leq & L(a_{j}-a_{j-1})^{\beta} \sum_{ {\tau}\in\mathcal{D}_{t,s}^j }   | \varphi_j(p_{\tau},x_{\tau})^{\top} (\Lambda_{t,s}^j)^{-1}  \varphi_j(p,x) | \\
   = & \frac{L(p_{max}-p_{min})^{\beta}}{N^{\beta}}\sum_{ {\tau}\in\mathcal{D}_{t,s}^j }  |  \varphi_j(p_{\tau},x_{\tau})^{\top} (\Lambda_{t,s}^j)^{-1}  \varphi_j(p,x)| .
   \end{align*}
   It remains to bound $\sum_{ {\tau}\in\mathcal{D}_{t,s}^j }  p | \varphi_j(p_{\tau},x_{\tau})^{\top} (\Lambda_{t,s}^j)^{-1}  \varphi_j(p,x) | $. 
   \begin{align*}
     & \sum_{ {\tau}\in\mathcal{D}_{t,s}^j } p |  \varphi_j(p_{\tau},x_{\tau})^{\top} (\Lambda_{t,s}^j)^{-1}  \varphi_j(p,x) |  \\
     \overset{(a)}{\leq} & p \sqrt{ |\mathcal{D}_{t,s}^j |   \sum_{ {\tau}\in\mathcal{D}_{t,s}^j }  |  \varphi_j(p_{\tau},x_{\tau})^{\top} (\Lambda_{t,s}^j)^{-1}  \varphi_j(p,x) |^2  } \\
     = & p \sqrt{ |\mathcal{D}_{t,s}^j |    \varphi_j(p,x)(\Lambda_{t,s}^j)^{-1}  \Big(\sum_{ {\tau}\in\mathcal{D}_{t,s}^j }\varphi_j(p_{\tau},x_{\tau})  \varphi_j(p_{\tau},x_{\tau})^{\top} \Big) (\Lambda_{t,s}^j)^{-1}  \varphi_j(p,x)   } \\
     \leq & p \sqrt{ |\mathcal{D}_{t,s}^j |    \varphi_j(p,x)  (\Lambda_{t,s}^j)^{-1}  \varphi_j(p,x)   } \\
     \overset{(b)}{\leq} & 2^{-s} p_{max} \sqrt{ |\mathcal{D}_{t,s}^j |     } \\
     \overset{(c)}{\leq} & p_{max}  \sqrt{ 2 (\alpha+\varpi(\beta)+1)\ln t  },
   \end{align*}
   where the inequality (a) is Cauchy-Schwarz inequality, 
   the inequality (b) is from the condition of the step (a) in \cref{alg: PADP},
   and the inequality (c) is due to \cref{lem: size of dataset j in linear contextual effect setting}.

   By the triangular inequality, we find 
   \begin{align*}
     & \Prob{ |p(\theta^j -\theta^j_{t,s})^{\top} \varphi_j(p,x) | \geq  \gamma p \sqrt{ \varphi_j(p,x)^{\top}  (\Lambda^j_{t,s})^{-1} \varphi_j(p,x) }  + \frac{L p_{max}(p_{max}-p_{min})^{\beta}  \sqrt{2 (\alpha+\varpi(\beta)+1)\ln t} }{ N^{\beta}  }   }  \\  
     \leq  & \mathbb{P} \Bigg( |  p (\theta_{t,s}^j -\tilde{\theta}^j_{t,s})^{\top} \varphi_j(p,x)  |     +  | p  (\theta^j -\tilde{\theta}^j_{t,s})^{\top} \varphi_j(p,x)  |  \\ 
     &  \quad \quad \quad \geq  \gamma p \sqrt{ \varphi_j(p,x)^{\top}  (\Lambda^j_{t,s})^{-1} \varphi_j(p,x) }  + \frac{Lp_{max}(p_{max}-p_{min})^{\beta}  \sqrt{2 (\alpha+\varpi(\beta)+1)\ln t} }{ N^{\beta}  } \Bigg)  \\
     \leq & \Prob{ |  (\theta^j -\tilde{\theta}^j_{t,s})^{\top} \varphi_j(p,x)  |   \geq  \gamma \sqrt{ \varphi_j(p,x)^{\top}  (\Lambda^j_{t,s})^{-1} \varphi_j(p,x) }   } \\
     \leq & \frac{\delta }{NS T}.
   \end{align*}

\end{myproof}

Based on \cref{lem: UCB in linear contextual effect setting}, we construct a high-probability event 
\begin{align*}
   \Gamma = &\Big\{ |p(\theta^j -\theta^j_{t,s})^{\top} \varphi_j(p,x) | \leq  \gamma p\sqrt{ \varphi_j(p,x)^{\top}  (\Lambda^j_{t,s})^{-1} \varphi_j(p,x) } \\ 
   & \quad \quad \quad + \frac{Lp_{max}(p_{max}-p_{min})^{\beta}  \sqrt{2 (\alpha+\varpi(\beta)+1)\ln t} }{ N^{\beta}  }, \forall t, \forall s,\forall j  \Big\},
\end{align*}
which satisfies 
$$
\prob{\overline{\Gamma}} \leq \frac{\delta  }{  NST } \times N \times T \times S = \delta .
$$
Similarly, we introduce the revenue function
$$
rev(p,x) = p(f(p)+\mu^\top x).
$$
Given the context $x_t$,
we define the index of the best price within the interval $[a_{j-1},a_j] $ at the candidate set $\mathcal{A}_{t,s}$ as 
$$
j_{t,s}^* = \argmax_{j\in \mathcal{A}_{t,s}} \sup_{p\in [a_{j-1},a_j] }  rev(p,x_t) .
$$
We also denote the context-dependent best price as $p_t^* \in \argmax_{p\in [p_{min},p_{max}] }  rev(p,x_t) $.

\begin{mylem}
 \label{lem: difference of best prices for each layer in linear contextual effect setting}
 Given the event $\Gamma$,
 then for each round $t$ and each layer $s\in [s_t-1]$, 
 we have 
 $$
  rev( p_t^* ,x_t  ) - rev( p_{t,s}^{j^*_{t,s}},x_t)   \leq   \frac{2Lp_{max}(p_{max}-p_{min})^{\beta}  \sqrt{2 (\alpha+\varpi(\beta)+1)\ln t} }{ N^{\beta}  }    (s-1).
 $$
\end{mylem}

\begin{myproof}
 We prove this lemma by induction on $s$. 
 For $s=1$, the lemma holds naturally as $\mathcal{A}_{t,1} = [N]$ and thus $rev( p_t^* ,x_t  )  =  rev( p_{t,1}^{j^*_{t,1}},x_t)  $.
 Assume that the bound holds in the layer $s$.
 It is sufficient to show that 
 $$ 
 rev( p^{j^*_{t,s}}_{t,s},x_t ) - rev( p^{j^*_{t,s+1}}_{t,s+1},x_t )   \leq  \frac{2Lp_{max}(p_{max}-p_{min})^{\beta}  \sqrt{2 (\alpha+\varpi(\beta)+1)\ln t} }{ N^{\beta}  } .
 $$
 If $j_{t,s}^* = j^*_{t,s+1}$, the desired bound holds.
 Hence we assume that $j^*_{t,s} \notin \mathcal{A}_{t,s+1}$.
 Let $\hat{j}_{t,s}:= \argmax_{j\in\mathcal{A}_{t,s}} \sup_{p\in [a_{j-1},a_j]} p U_{t,s}^j (p) $ be the index with the highest UCB in $\mathcal{A}_{t,s}$.
 From the step (a) we know $\hat{j}_{t,s} \in \mathcal{A}_{t,s+1}.$ Then we have 
 \begin{align*}
   & rev( p^{j^*_{t,s}}_{t,s},x_t ) - rev( p^{j^*_{t,s+1}}_{t,s+1},x_t )  \\
  \leq & rev( p^{j^*_{t,s}}_{t,s},x_t ) - rev( p^{\hat{j}_{t,s+1}}_{t,s+1} ,x_t)   \\
  \leq &  p^{j^*_{t,s}}_{t,s} \varphi_{ j^*_{t,s}  } ( p^{j^*_{t,s}}_{t,s},x_t )^{\top}\theta^{j^*_{t,s}}  -p^{\hat{j}_{t,s}}_{t,s} \varphi_{ \hat{j}_{t,s}  } ( p^{\hat{j}_{t,s}}_{t,s},x_t )^{\top}\theta^{\hat{j}_{t,s}}  + \frac{2 p_{max} L(p_{max}-p_{min})^{\beta} } { N^{\beta}  } ,
 \end{align*}
 where the last inequality is due to \cref{lem: polynomial regression}.
 From the definition of $\Gamma$, we know that 
 \begin{align*}
  & p^{j^*_{t,s}}_{t,s} \varphi_{ j^*_{t,s}  } ( p^{j^*_{t,s}}_{t,s},x_t )^{\top}\theta^{j^*_{t,s}}  -p^{\hat{j}_{t,s}}_{t,s} \varphi_{ \hat{j}_{t,s}  } ( p^{\hat{j}_{t,s}}_{t,s},x_t )^{\top}\theta^{\hat{j}_{t,s}}     \\ 
 \leq & p^{j^*_{t,s}}_{t,s} \varphi_{ j^*_{t,s}  } ( p^{j^*_{t,s}}_{t,s},x_t )^{\top}\theta_{t,s}^{j^*_{t,s}}+ \gamma p^{j^*_{t,s}}_{t,s} \sqrt{ \varphi_{j^*_{t,s}}(p^{j^*_{t,s}}_{t,s},x_t)^{\top}  (\Lambda^{j^*_{t,s}}_{t,s})^{-1} \varphi_{j^*_{t,s}}(p^{j^*_{t,s}}_{t,s},x_t) }  \\ 
 & \quad \quad    -   p^{\hat{j}_{t,s}}_{t,s} \varphi_{ \hat{j}_{t,s}  } ( p^{\hat{j}_{t,s}}_{t,s},x_t )^{\top}\theta_{t,s}^{\hat{j}_{t,s}}  +  \gamma p^{\hat{j}_{t,s}}_{t,s}\sqrt{ \varphi_{\hat{j}_{t,s}}(p^{\hat{j}_{t,s}}_{t,s},x_t)^{\top}  (\Lambda^{\hat{j}_{t,s}}_{t,s} )^{-1} \varphi_{\hat{j}_{t,s}}(p^{\hat{j}_{t,s}}_{t,s},x_t) }   \\ 
 & \quad \quad\quad \quad + \frac{2Lp_{max}(p_{max}-p_{min})^{\beta}  \sqrt{2 (\alpha+\varpi(\beta)+1)\ln t} }{ N^{\beta}  } \\
 = &  p^{j^*_{t,s}}_{t,s} \varphi_{ j^*_{t,s}  } ( p^{j^*_{t,s}}_{t,s},x_t )^{\top}\theta_{t,s}^{j^*_{t,s}}+ \gamma p^{j^*_{t,s}}_{t,s} \sqrt{ \varphi_{j^*_{t,s}}(p^{j^*_{t,s}}_{t,s},x_t)^{\top}  (\Lambda^{j^*_{t,s}}_{t,s})^{-1} \varphi_{j^*_{t,s}}(p^{j^*_{t,s}}_{t,s}) }      \\ 
 & \quad \quad -   p^{\hat{j}_{t,s}}_{t,s} \varphi_{ \hat{j}_{t,s}  } ( p^{\hat{j}_{t,s}}_{t,s},x_t )^{\top}\theta_{t,s}^{\hat{j}_{t,s}}  - \gamma p^{\hat{j}_{t,s}}_{t,s}\sqrt{ \varphi_{\hat{j}_{t,s}}(p^{\hat{j}_{t,s}}_{t,s},x_t)^{\top}  (\Lambda^{\hat{j}_{t,s}}_{t,s} )^{-1} \varphi_{\hat{j}_{t,s}}(p^{\hat{j}_{t,s}}_{t,s},x_t) }   \\ 
 & \quad \quad\quad \quad + 2\gamma p^{\hat{j}_{t,s}}_{t,s}\sqrt{ \varphi_{\hat{j}_{t,s}}(p^{\hat{j}_{t,s}}_{t,s},x_t)^{\top}  (\Lambda^{\hat{j}_{t,s}}_{t,s} )^{-1} \varphi_{\hat{j}_{t,s}}(p^{\hat{j}_{t,s}}_{t,s},x_t) }   + \frac{2Lp_{max}(p_{max}-p_{min})^{\beta}  \sqrt{2 (\alpha+\varpi(\beta)+1)\ln t} }{ N^{\beta}  }\\
=  & p^{j^*_{t,s}}_{t,s}  U_{t,s}^{j^*_{t,s}}( p^{j^*_{t,s}}_{t,s} )  - p^{\hat{j}_{t,s}}_{t,s}   U_{t,s}^{\hat{j}_{t,s}}( p^{\hat{j}_{t,s}}_{t,s}  )   \\ 
& \quad \quad + 2\gamma p^{\hat{j}_{t,s}}_{t,s}\sqrt{ \varphi_{\hat{j}_{t,s}}(p^{\hat{j}_{t,s}}_{t,s},x_t)^{\top}  (\Lambda^{\hat{j}_{t,s}}_{t,s} )^{-1} \varphi_{\hat{j}_{t,s}}(p^{\hat{j}_{t,s}}_{t,s},x_t) }   + \frac{2Lp_{max}(p_{max}-p_{min})^{\beta}  \sqrt{2 (\alpha+\varpi(\beta)+1)\ln t} }{ N^{\beta}  }.
\end{align*}

From the algorithmic construction (step (b)), we know 
$$
\gamma p^{\hat{j}_{t,s}}_{t,s}\sqrt{ \varphi_{\hat{j}_{t,s}}(p^{\hat{j}_{t,s}}_{t,s},x_t)^{\top}  (\Lambda^{\hat{j}_{t,s}}_{t,s} )^{-1} \varphi_{\hat{j}_{t,s}}(p^{\hat{j}_{t,s}}_{t,s},x_t) }  \leq p_{max}2^{-s}.
$$
Since $ {j}^*_{t,s} \notin \mathcal{A}_{t,s} $ by our assumption, the step (a) yields 
\begin{align*}
 & p^{\hat{j}_{t,s}}_{t,s}   U_{t,s}^{\hat{j}_{t,s}}( p^{\hat{j}_{t,s}}_{t,s}  )   - p^{j^*_{t,s}}_{t,s}  U_{t,s}^{j^*_{t,s}}( p^{j^*_{t,s}}_{t,s} )\\
\geq &  p^{\hat{j}_{t,s}}_{t,s}   U_{t,s}^{\hat{j}_{t,s}}( p^{\hat{j}_{t,s}}_{t,s}  )   -   \sup_{p\in [a_{j^*_{t,s} -1}, a_{j^*_{t,s} }]} pU_{t,s}^{j^*_{t,s}}(p) \\
 >  & p_{max}2^{1-s} \\ 
\geq & 2 \gamma p^{\hat{j}_{t,s}}_{t,s}\sqrt{ \varphi_j(p^{\hat{j}_{t,s}}_{t,s},x_t)^{\top}  \Lambda^j_{t,s} \varphi_j(p^{\hat{j}_{t,s}}_{t,s},x_t) }.
\end{align*}

Combing all inequalities above, we obtain 
\begin{align*}
 rev( p^{j^*_{t,s}}_{t,s},x_t ) - rev( p^{j^*_{t,s+1}}_{t,s+1},x_t )   \leq  \frac{2Lp_{max}(p_{max}-p_{min})^{\beta}  \sqrt{2 (\alpha+\varpi(\beta)+1)\ln t} }{ N^{\beta}  } .
\end{align*}

\end{myproof}

\begin{mylem}
 \label{lem: regret of prices at layer s in linear contextual effect setting}
 Given the event $\Gamma$,
 then for each round $t$, 
 we have 
   \begin{align*}
     p^{j^*_{t,s}}_{t,s} \varphi_{ j^*_{t,s}  } ( p^{j^*_{t,s}}_{t,s},x_t )^{\top}\theta^{j^*_{t,s}}   -   p_t \varphi_{j_t}(p_t,x_t)^{\top} \theta^{ j_t  }   \leq  6 p_{max} \cdot 2^{-s} +   \frac{Lp_{max}(p_{max}-p_{min})^{\beta}  \sqrt{2 (\alpha+\varpi(\beta)+1)\ln t} }{ N^{\beta}  } ,
   \end{align*}
 for all $2\leq s < s_t$.
\end{mylem}
\begin{myproof}
   For any layer $s < s_t$, the step (a) shows that 
   \begin{align*}
     p_{max} 2^{1-s} \geq  & \max \limits_{  j' \in \mathcal{A}_{t,s} } \sup_{p\in [a_{j-1},a_j] } p U^{j'}_{t,s}(p)   -  \sup_{ p \in [a_{j_t-1},a_{j_t}] } p  U^{j_t}_{t,s}(p)  \\
     \geq   &  p^{j^*_{t,s}}_{t,s} U^{j^*_{t,s}}_{t,s}(p^{j^*_{t,s}}_{t,s})   -  p_t \varphi_{j_t}(p_t,x_t)^{\top} \theta^{ j_t  }  +  \frac{Lp_{max}(p_{max}-p_{min})^{\beta}  \sqrt{2 (\alpha+\varpi(\beta)+1)\ln t} }{ N^{\beta}  },
   \end{align*}
   where the last inequality is due to the definition of $\Gamma$.
   
   The step (b) implies that 
   $$
   \sup_{  p \in [a_{j_t-1},a_{j_t}]  }  \gamma p  \sqrt{ \varphi_{j_t}(p,x_t )^{\top}  ( \Lambda^{j_t}_{t,s})^{-1}\varphi_{j_t}(p,x_t ) } \leq   p_{max}2^{1-s}
   $$
   as $j_t\in\mathcal{A}_{t,s}$ and $s<s_t$. 
   Combing two inequalities we obtain 
   \begin{align*}
   &  p^{j^*_{t,s}}_{t,s} U_{ j^*_{t,s}  } ( p^{j^*_{t,s}}_{t,s} )   -\sup_{p \in [a_{j_{t}-1},a_{j_t}]}  p \bigg(U_{t,s}^{j_t}(p) -   2\gamma   \sqrt{ \varphi_{j_t}(p)^{\top}  ( \Lambda^{j_t}_{t,s})^{-1}\varphi_{j_t}(p) } \bigg)   \\
   \leq & 2 \sup_{p \in [a_{j_{t}-1},a_{j_t}]} \gamma p  \sqrt{ \varphi_{j_t}(p,x_t)^{\top}  ( \Lambda^{j_t}_{t,s})^{-1}\varphi_{j_t}(p,x_t) }  +  p_{max} 2^{1-s} \\
   \leq & 3 p_{max} 2^{1-s}.
   \end{align*}

   Therefore, from the definition of $\Gamma$ we have 
   \begin{align*}
      &  3 p_{max} 2^{1-s}    \\
   \geq & p^{j^*_{t,s}}_{t,s} U_{ j^*_{t,s}  } ( p^{j^*_{t,s}}_{t,s} )   -\sup_{p \in [a_{j_{t}-1},a_{j_t}]}  p \bigg(U_{t,s}^{j_t}(p) -   2\gamma   \sqrt{ \varphi_{j_t}(p,x_t)^{\top}  ( \Lambda^{j_t}_{t,s})^{-1}\varphi_{j_t}(p,x_t) } \bigg)   \\
   \geq & p^{j^*_{t,s}}_{t,s} \varphi_{ j^*_{t,s}  } ( p^{j^*_{t,s}}_{t,s},x_t )^{\top}\theta^{j^*_{t,s}}   - \sup_{p\in[a_{j_t-1},a_{j_t}]} p \varphi_{j_t}(p,x_t)^{\top} \theta^{ j_t  }     - \frac{2Lp_{max}(p_{max}-p_{min})^{\beta}  \sqrt{2 (\alpha+\varpi(\beta)+1)\ln t} }{ N^{\beta}  }   .
   \end{align*}
  Rearranging all terms yields the desired inequality.

\end{myproof}

\begin{mylem}
 \label{lem: discrete regret at the round t in linear contextual effect setting}
 Given the event $\Gamma$, for every round $t$, we have 
 $$
 rev( p_t^* ,x_t  ) - rev( p_t ,x_t) \leq 12 p_{max} \cdot  2^{-s_t} + \frac{Lp_{max}(p_{max}-p_{min})^{\beta}  \sqrt{2 (\alpha+\varpi(\beta)+1)\ln t} }{ N^{\beta}  }    (2s_t-1).
 $$
\end{mylem}

\begin{myproof}
 It follows from \cref{lem: polynomial regression} and \cref{lem: regret of prices at layer s in linear contextual effect setting} that 
 \begin{align*}
  & rev( p_t^*,x_t   ) - rev( p_t ,x_t) \\
  \leq &   rev( p_t^*,x_t  ) - rev( p_{t,s_t-1}^{j^*_{t,s_t-1}}, x_t)  +  rev( p_{t,s_t-1}^{j^*_{t,s_t-1}},x_t) -  rev( p_t,x_t )  \\
  \leq &  \frac{2 Lp_{max}(p_{max}-p_{min})^{\beta}  \sqrt{2 (\alpha+\varpi(\beta)+1)\ln t} }{ N^{\beta}  }    (s_t-2)+  rev( p_{t,s_t-1}^{j^*_{t,s_t-1}},x_t) -  rev( p_t,x_t )  \\
  \leq & \frac{Lp_{max}(p_{max}-p_{min})^{\beta}  \sqrt{2 (\alpha+\varpi(\beta)+1)\ln t} }{ N^{\beta}  }    (2s_t-2) \\
  & \quad \quad \quad +  p^{j^*_{t,s_t-1}}_{t,s_t-1} \varphi_{ j^*_{t,s_t-1}  } ( p^{j^*_{t,s_t-1}}_{t,s_t-1},x_t )^{\top}\theta^{j^*_{t,s_t-1}}  -  p_t \varphi_{ j_t  } ( p_t,x_t )^{\top}\theta^{j_t}  \\
  \leq & \frac{Lp_{max}(p_{max}-p_{min})^{\beta}  \sqrt{2 (\alpha+\varpi(\beta)+1)\ln t} }{ N^{\beta}  }    (2s_t-2) \\ 
  & \quad \quad \quad + 6p_{max} \cdot 2^{1-s_t} +   \frac{Lp_{max}(p_{max}-p_{min})^{\beta}  \sqrt{2 (\alpha+\varpi(\beta)+1)\ln t} }{ N^{\beta}  } \\
  \leq& 12 p_{max} \cdot   2^{-s_t}+ \frac{Lp_{max}(p_{max}-p_{min})^{\beta}  \sqrt{2 (\alpha+\varpi(\beta)+1)\ln t} }{ N^{\beta}  }    (2s_t-1).
 \end{align*}

\end{myproof}

\begin{mylem}
   \label{lem: bound of r in linear contextual effect setting}
   Assuming $|\mathcal{D}_{t,s}^j|\geq 2$, then for arbitrary $t$ and $s$, we have
   $$
   \sum_{\tau\in \mathcal{D}_{t,s}^j} p_{\tau}  \gamma  \sqrt{ \varphi_j(p_{\tau},x_{\tau})^{\top}  (\Lambda^j_{\tau,s})^{-1} \varphi_j(p_{\tau},x_{\tau}) }     \leq \gamma p_{max} \sqrt{2 (\alpha+\varpi(\beta)+1) | \mathcal{D}_{t,s}^j| \ln | \mathcal{D}_{t,s}^j|  }.
   $$
\end{mylem}

\begin{myproof}
   We have
   $$
   \sum_{\tau\in \mathcal{D}_{t,s}^j} p_{\tau}  \gamma  \sqrt{ \varphi_j(p_{\tau},x_{\tau})^{\top}  (\Lambda^j_{\tau,s})^{-1} \varphi_j(p_{\tau},x_{\tau}) } \leq  \gamma   p_{max}  \sum_{\tau\in \mathcal{D}_{t,s}^j}    \sqrt{ \varphi_j(p_{\tau},x_{\tau})^{\top}  (\Lambda^j_{\tau,s})^{-1} \varphi_j(p_{\tau},x_{\tau}) }
   $$
   From the lemma 13 in \citep{supLinUCB} or the lemma 3 in \citep{linContextual}, we know that 
   $$
   \sum_{\tau\in \mathcal{D}_{t,s}^j}    \sqrt{ \varphi_j(p_{\tau},x_{\tau})^{\top}  (\Lambda^j_{\tau,s})^{-1} \varphi_j(p_{\tau},x_{\tau}) } \leq \sqrt{2 (\alpha+\varpi(\beta)+1) | \mathcal{D}_{t,s}^j| \ln | \mathcal{D}_{t,s}^j|  }
   $$

\end{myproof}

\begin{mylem}
   \label{lem: size of dataset j in linear contextual effect setting}
   For all $s$, we have 
   $$
   |\mathcal{D}_{t,s}^j | \leq 2^s \gamma \sqrt{2 (\alpha+\varpi(\beta)+1) | \mathcal{D}_{t,s}^j| \ln | \mathcal{D}_{t,s}^j|  }.
   $$
\end{mylem}

\begin{myproof}
   From the step (b), we know that 
   $$
   \sum_{\tau\in \mathcal{D}_{t,s}^j} p_{\tau}  \gamma  \sqrt{ \varphi_j(p_{\tau},x_{\tau})^{\top}  (\Lambda^j_{\tau,s})^{-1} \varphi_j(p_{\tau},x_{\tau}) }    \geq p_{max} 2^{-s} |\mathcal{D}_{t,s}^j |.
   $$
   By \cref{lem: bound of r}, we obtain 
   $$
   \sum_{\tau\in \mathcal{D}_{t,s}^j} p_{\tau}  \gamma  \sqrt{ \varphi_j(p_{\tau},x_{\tau})^{\top}  (\Lambda^j_{\tau,s})^{-1} \varphi_j(p_{\tau},x_{\tau}) }    \leq  \gamma p_{max} \sqrt{2 (\alpha+\varpi(\beta)+1) | \mathcal{D}_{t,s}^j| \ln | \mathcal{D}_{t,s}^j|  }.
   $$
   Therefore, combing above inequalities, we have 
   $$
   |\mathcal{D}_{t,s}^j | \leq 2^s \gamma \sqrt{2 (\alpha+\varpi(\beta)+1) | \mathcal{D}_{t,s}^j| \ln | \mathcal{D}_{t,s}^j|  }.
   $$

\end{myproof}

Now we have all materials to prove \cref{thm: regret upper bound of contextual dynamic pricing algorithm}.

\begin{myproof}
   Let $\Psi_0$ be the set of rounds for which an alternative is chosen when $   p_t \sqrt{ \phi_j(p_t,x_t )^{\top}  (\Lambda^j_{t,1})^{-1} \phi_j(p_t,x_t ) }     \leq 1/\sqrt{T}$.
   Since $2^{-S} \leq 1/\sqrt{T}$, we have $ \Psi_0 \cup \cup_{s\in [S]}\Psi^s_{T+1} = [T]$. 
   
   Recall that $p_t^*$ maximizes $rev(p,x_t)$.
    \begin{align*}
       & \mathbb{E}[  Reg(T) ] \\
       = &   \sum_{t=1   }^{T}  \mathbb{E}\left[ rev(p_t^*,x_t)- rev(p_t,x_t)   \right] \\ 
       = & \sum_{s=2}^S \sum_{t\in \Psi_{T +1}^s } \mathbb{E}\Big[   rev(p_t^*,x_t)- rev(p_t,x_t)  \Big] + \sum_{t\in \Psi_{T +1}^1 } \mathbb{E}\Big[  rev(p_t^*,x_t)- rev(p_t,x_t)  \Big] \\ 
         & \quad\quad + \sum_{t\in \Psi_0 } \mathbb{E}\Big[  rev(p_t^*,x_t)- rev(p_t,x_t)  \Big] \\ 
     \leq&  \sum_{s=2}^S \left( 12 p_{max} \cdot 2^{-s}+ \frac{Lp_{max}(p_{max}-p_{min})^{\beta}  \sqrt{2 (\alpha+\varpi(\beta) +1)\ln T} }{ N^{\beta}  }    (2s-1) \right)  |\Psi_{T+1}^s| \\
     &  \quad \quad\quad\quad +p_{max}|\Psi_{T+1}^1|    +  \sum_{t\in \Psi_0 } \mathbb{E}\Big[ rev(p_t^*,x_t)- rev(p_t,x_t)    \Big]     ,
   \end{align*}
   where the last inequality is due to \cref{lem: discrete regret at the round t in linear contextual effect setting}.

  From \cref{lem: size of dataset j in linear contextual effect setting}, we have 
  $$
  |\Psi_{T +1}^s| \leq 2^s \gamma \sqrt{2N (\alpha+\varpi(\beta)+1 )  | \Psi_{T+1}^s| \ln T },
  $$
  which implies 
  $$
  |\Psi_{T +1}^s| \leq 2^{2s+1} \gamma^2 N (\varpi(\beta) +1) \ln T.
  $$
  As a corollary, we have 
  $$
  p_{max}|\Psi_{T+1}^1| \leq 8\gamma^2  p_{max} N (\varpi(\beta) +1) \ln^2 (T/\delta).
  $$

  Therefore, we have 
  \begin{align*}
   \sum_{s=2}^S 12\gamma p_{max} \cdot 2^{-s}|\Psi_{T +1}^s| \leq & \sum_{s=2}^S  12p_{max}  \sqrt{2N (\varpi(\beta) +1) | \Psi_{T+1}^s| \ln T } \\
  & \leq  12 \gamma p_{max}\sqrt{2N (\varpi(\beta) +1)(S-1)\sum_{s=2}^S  | \Psi_{T+1}^s| \ln T } \\
  & \leq  12 \gamma p_{max}\sqrt{N (\varpi(\beta) +1) {T}  \ln T  \log_2 T    }
   \end{align*}
  and
  \begin{align*}
    & \sum_{s=2}^S   \frac{Lp_{max}(p_{max}-p_{min})^{\beta}  \sqrt{2 (\varpi(\beta) +1) \ln T} }{ N^{\beta}  }    (2s-1)   |\Psi_{T+1}^s| \\ 
   \leq & \frac{2 T Lp_{max}(p_{max}-p_{min})^{\beta}  \sqrt{2 (\varpi(\beta) +1) \ln T} }{ N^{\beta}  }   \log_2 T. 
  \end{align*}
  
  From \cref{lem: regret at the round t in Psi0}, we know that 
 $$
 \sum_{t\in \Psi_0 } \mathbb{E}\Big[   p^*f(p^*) - p_tf(p_t)    \Big] \leq 2p_{max} \sqrt{T} +  \frac{2Lp_{max}(p_{max}-p_{min})^{\beta}  }{ N^{\beta}  } .
 $$

 For the regret caused in the initialization step,
 we directly bound the caused regret as 
 $
   p_{max}\mathcal{T}.
 $
 Thanks to \cref{lem: stopping time}, we have 
 $$
 p_{max}\mathbb{E}[\mathcal{T}] \leq C_0 p_{max}N \log T
 $$

   Since $N = \lceil  (p_{max}-p_{min})^{\frac{2\beta}{2\beta+1}}   T^{\frac{1}{2\beta+1}}     \rceil$, we have 
   \begin{align*}
     & \mathbb{E}[Reg(T)] \leq (12+  L  )p_{max} (p_{max}-p_{min})^{\frac{\beta}{2\beta+1}}   (\varpi(\beta) +1)^{\frac{1}{2 }}  \log^{\frac{3}{2}}_2 (T/\delta ) T^{ \frac{\beta+1}{2\beta+1} }  \\ 
     & \quad \quad \quad +  2p_{max} \sqrt{T} + (17+C_0) p_{max}  (p_{max}-p_{min})^{\frac{2\beta}{2\beta+1}} (\varpi(\beta) +1)^{\frac{3}{2 }} ( T \log_2 T )^{\frac{1}{2\beta+1}}     \ln T .
\end{align*}
\end{myproof}

\section{Related Material}

\begin{mylem}(\citep{Multimodal_DP})
   \label{lem: polynomial regression}
   Suppose $f$ satisfies \cref{asp: Smoothness} and let $[a,b] \subset [p_{min},p_{max}]$ be an arbitrary interval. 
   There exists a $\varpi (\beta)$-degree polynomial $P_{[a,b]}(x)$ such that
   $$
   \sup_{x\in [a,b]}|f(x) - P_{[a,b]}(x)| \leq L (b-a)^{\beta}
   $$
 \end{mylem} 

\begin{mylem}(\citep{GLbandit})
   \label{lem: eigenvalue}
   Define $V_n = \sum_{t=1}^n x_tx_t^{\top}$, where $x_t$ is drawn i.i.d. from some distribution with support in the $d$-dimensional unit ball.
   Futhermore, let $\Sigma = \mathbb{E}[xx^{\top}]$ be thr Gram matrix, and $B,\delta$ be two positive constants.
   Then, there exist positive, universal constants $C_1$ and $C_2$ such that $\lambda_{min}(V_n)\geq B$ with probability at least $1-\delta$,
   as long as 
   $$
   n \geq \left( \frac{C_1\sqrt{d}+C_2\sqrt{\log(1/\delta)}}{\lambda_{min}(\Sigma)}    \right)^2 + \frac{2B}{\lambda_{min}(\Sigma)}.
   $$
\end{mylem}

\begin{mylem}
   \label{lem: concentration of beta}
   With an upper bound $\beta_{max}$ of the smoothness parameter and \cref{asp: Smoothness}, 
   then for some constant $C>0$, we have 
   $$
   \hat{\beta} \in [\beta - \frac{4(\beta_{max}+1) \ln \ln T  }{ \ln T  },\beta   ]
   $$
   with probability at least $1- e^{-C \ln^2 T} $.
\end{mylem}


\end{document}